\documentclass[journal]{IEEEtran}

\usepackage{cite}
\usepackage{graphicx}
\usepackage{amsmath}
\usepackage{amssymb}
\usepackage{amsthm}
\usepackage{subcaption}
\usepackage{float}
\usepackage{bbm}

\usepackage{tabularx}
\usepackage{booktabs}

\theoremstyle{plain}

\theoremstyle{definition}

\theoremstyle{remark}

\begin{document}
\title{Clarity-Driven Adaptive Mission Planning for a Solar-Powered Autonomous Surface Vessel}
\author{Kavin M. Govindarajan, Kaleb Ben Naveed, David Li, Devansh R. Agrawal, Mary Maceda, Matthew Bryant, Dimitra Panagou, Chris Vermillion%
\thanks{This work has been submitted to the IEEE for possible publication. Copyright may be transferred without notice, after which this version may no longer be accessible. This work was supported by the National Science Foundation under Awards 2223844 and 2223845. \textit{(Corresponding author: Kavin M. Govindarajan.)}}
\thanks{Kavin M. Govindarajan and Kaleb Ben Naveed are with the Department of Robotics, University of Michigan, Ann Arbor, MI 48109, USA (e-mail: kmgovind@umich.edu; kbnaveed@umich.edu).}%
\thanks{David Li is with the Department of Mechanical Engineering, University of Michigan, Ann Arbor, MI 48109, USA (e-mail: davidlxl@umich.edu).}%
\thanks{Devansh R. Agrawal is a graduate of the Department of Aerospace Engineering, University of Michigan, Ann Arbor, MI 48109, USA (e-mail: devansh@umich.edu).}%
\thanks{Mary Maceda is with the Department of Mechanical and Aerospace Engineering, North Carolina State University, Raleigh, NC 27695, USA (e-mail: memaceda@ncsu.edu).}%
\thanks{Matthew Bryant is with the Department of Mechanical and Aerospace Engineering, North Carolina State University, Raleigh, NC 27695, USA (e-mail: mbryant@ncsu.edu).}%
\thanks{Dimitra Panagou is with the Department of Robotics, University of Michigan, Ann Arbor, MI 48109, USA (e-mail: dpanagou@umich.edu).}%
\thanks{Chris Vermillion is with the Department of Mechanical Engineering, University of Michigan, Ann Arbor, MI 48109, USA (e-mail: cvermill@umich.edu).}%
}

\maketitle
\thispagestyle{plain}
\pagestyle{plain}

\begin{abstract}
This paper presents a comprehensive methodology, backed by simulation and experimental results, for an adaptive mission planning strategy that is mathematically grounded in the maximization of the information-driven quantity of \textit{clarity}. In particular, the methodology introduced in this work addresses the use of sustainably powered autonomous surface vessels (ASVs) to perform scientific missions that prioritize locations of maximum interest. Because these locations of maximum interest are not known a priori but are rather identified based on in-situ measurements, online adaptation is essential for mission planning. Furthermore, the reliance on a limited-power renewable resource requires an adaptation strategy that maximizes information gathering subject to mobility restrictions imposed by the renewable resource. Our approach addresses these challenges through a clarity-driven ergodic search, where clarity represents a mathematical quantity that has been rigorously linked to differential (information) entropy. In our approach, a target clarity map is adapted online to reflect the prioritization of certain areas based on in-situ measurements, and an ergodic path planner continually re-plans based on the target clarity map and ASV speed limitations (driven by the renewable propulsion source). Focusing on a solar-powered ASV, we present simulations and experiments conducted at Jordan Lake, North Carolina, demonstrating the effectiveness of the proposed approach.
\end{abstract}


\section{Introduction}\label{sec:intro}
\IEEEPARstart{T}{he} collection of spatiotemporally varying data is critical for many environmental and engineering applications: characterizing wind and ocean currents enables the identification of energy-harvesting sites~\cite{naik2024integrated}, surface temperature data supports improved weather forecasting~\cite{huda2023machine}, and salinity data enhances understanding of climate-driven processes~\cite{seim2022overview}. While fixed platforms (e.g., moored sensor arrays~\cite{adcp} and high-frequency radar~\cite{radar}) enable data acquisition over extended periods, there has been recent emphasis on using \textbf{\textit{mobile}} observational platforms that can adjust their physical locations to target areas of interest, which are often unknown a priori and can change over time. Mobile observation can be accomplished through vessels equipped with finite amounts of energy (e.g., diesel-powered research vessels~\cite{muglia_transect} and undersea gliders~\cite{glider,glider2}) or through renewably powered platforms (e.g., sailing drones, wave gliders, or solar-powered autonomous surface vessels (ASVs)) that enable longer-term missions. Example applications include tracking plumes of pollutant or contaminant dispersion~\cite{zhang2011estimation,petillo2012autonomous}, adaptive front-following for estimating transport and mixing processes~\cite{petillo2015autonomous,mcdonald2020navigation}, and resource measurement (e.g., wind or ocean current) for siting energy-harvesting devices~\cite{changSiteSelectionOcean2015,gkeka-serpetsidakiReviewSustainableSiting2024}.

Regardless of the motivation for adaptation or whether the mobile observational platform at hand is powered by a finite or renewable resource, the basic objective is the same: \textbf{\textit{maximize the collection of useful scientific information under a limited energy resource}}. In the case of a finite energy resource, the limitations map to a restricted range that is influenced by the device's velocity profile. In the case of a renewable energy resource, the total range is not limited, but the achievable velocity profile is constrained by the available power. In either case, both the energy and information resources evolve over time. Notably, the information known about the scientific resource at any given location can increase due to sensor measurements but can also decay if that point goes unvisited for a long time. Ultimately, therefore, a unified framework for adaptive mission planning should satisfy three features:
\begin{enumerate}
\item Rigorous consideration of an evolving information field (where information is gained through observation but will also decay over time at points that go unvisited), based on observations taken;
\item Explicit consideration of an evolving battery state of charge, based on the environmental (solar) energy input and energy used for mobility;
\item Real-time adaptation based on in-situ scientific measurements.
\end{enumerate}

Information-driven planning has been heavily popularized (see~\cite{bai2021information, popovic2024learning}), with efforts often tying information measures to statistical quantities like entropy. To capture complex spatial fields, recent robotic information gathering frameworks employ nonstationary spatiotemporal models for autonomous surface vehicles~\cite{chen2022attentivekernel}. However, energy constraints in such formulations are typically imposed as simplistic ``budgets''~\cite{energybudget} (e.g., total distance, time, or control effort as a surrogate~\cite{emam2021data}) without accounting for vehicle power dynamics or state of charge (SOC)~\cite{notomista2021online}. Furthermore, standard IPP formulations generally assume non-decaying information, optimizing over truncated horizons for one-off coverage. Consequently, these approaches are prone to local extrema and struggle to sustain long-term tracking of dynamic scientific resources where unvisited locations decay in certainty over time.

A notable exception addressing this limitation is the information-theoretic quantity \textbf{\textit{clarity}}~\cite{agrawalSensorBasedPlanningControl2023}, derived from differential entropy~\cite{cover1999elements}, which explicitly models time-varying certainty decay. A natural planning strategy for clarity is \textbf{\textit{ergodic search}}~\cite{mathew2011metrics}. By matching the Fourier coefficients of the vehicle's trajectory statistics to a target spatial distribution (TSD), ergodic control inherently balances exploration and exploitation over long horizons without heuristic switching. Recent efforts have begun incorporating energy awareness into ergodic control: Seewald et al.~\cite{seewald2024energyawareergodicsearch} integrated battery models for multi-agent coverage, while Naveed et al.~\cite{naveed2024eclares} introduced clarity-driven ergodic search under SOC constraints. However, existing energy-aware ergodic methods rely on static target distributions defined offline, or achieve persistence through multi-agent duty cycling to fixed charging stations rather than optimizing trajectory-level energy dynamics for continuous operation of a single sustainably powered vehicle. In contrast, our work enables real-time, online adaptation of the target clarity map based on in-situ scientific measurements, allowing the planner to dynamically re-prioritize informative regions while explicitly controlling speed to satisfy energy constraints.

The present work extends the framework of~\cite{naveed2024eclares} by (i) enabling online adaptation according to in-situ measurements, (ii) augmenting the ergodic adaptive mission planner with energy-informed speed control, and (iii) validating the integrated framework experimentally. In particular, we utilize a \textbf{\textit{spatiotemporal Gaussian Process Kalman Filter (STGPKF)}} to maintain real-time estimates of the prediction means and variances of the scientific quantity of interest, based on real-time, in-situ measurements. From the prediction mean and variance models, we generate both an \textbf{\textit{actual clarity map}} and a \textbf{\textit{target clarity map}}. These maps jointly define the ergodic objective, allowing the controller to prioritize scientifically meaningful regions as the environment evolves. Additionally, following our earlier work~\cite{govindarajan2023} that established the optimality of piecewise constant speeds when the battery state of charge did not satisfy tightened saturation limits at equality, this speed control approach is incorporated into the ergodic planning problem.

\begin{figure}
\centering
\includegraphics[width=\columnwidth]{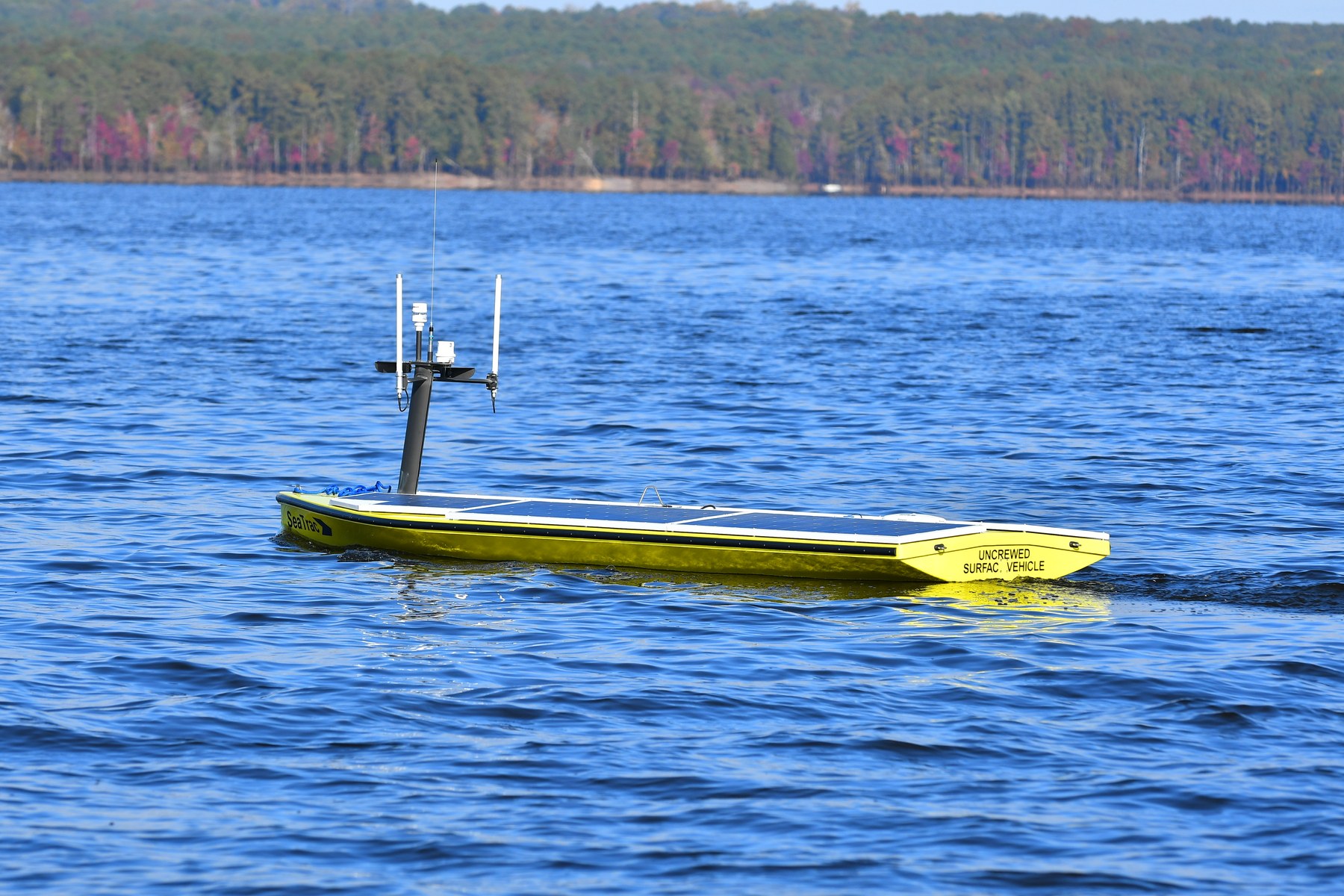}
\caption{SeaTrac ASV considered in this work~\cite{seatrac}.}
\label{fig:seatrac-real}
\end{figure}

To ground the problem in a real deployment scenario, we focus on a solar-powered ASV, namely the SeaTrac SP-48 (shown in Fig.~\ref{fig:seatrac-real}). Both simulation and field trials consider a mission domain in Jordan Lake, North Carolina, where the scientific objective is to characterize the evolving wind resource. Motivated by wind turbine siting studies, wherein regions with wind speeds close to a \textbf{\textit{rated}} value are of primary interest, we construct the target clarity map from the STGPKF wind speed estimates and their deviation from the rated speed.

In summary, the contributions of this paper are as follows:
\begin{enumerate}
    \item We develop an \textbf{\textit{energy-aware, adaptive clarity-driven ergodic control framework}} that unifies a statistically rigorous information metric with energy-aware speed control.
    \item We introduce an STGPKF-based method that generates real-time actual and target clarity maps from online measurements, enabling fully adaptive, science-aware planning.
    \item We validate the complete framework using both simulations and field experiments on the SeaTrac SP-48 ASV at Jordan Lake, demonstrating that the system actively targets informative regions while respecting energy and speed constraints.
\end{enumerate}

The remainder of the paper is organized as follows. Section~\ref{sec:prelim} introduces the information (clarity) and energy models used throughout this work. Section~\ref{sec:methods} presents the problem formulation and proposed control framework. Section~\ref{sec:simulation} evaluates performance in simulation. Section~\ref{sec:experiment} presents field results from a wind characterization mission on the SeaTrac ASV. Section~\ref{sec:conclusion} concludes and outlines future research directions.

\section{Energy \& Information (Clarity) Modeling}\label{sec:prelim}
The ultimate goal of this work is to present a process for maximizing the collection of useful scientific \textbf{\textit{information}} under a limited \textbf{\textit{energy}} resource. Therefore, it is critical to establish underlying models that describe how both energy and information (to be modeled using clarity) evolve over time. Those models are presented in this section.

\subsection{Energy Dynamics}\label{sec:prelim-energydynamics}
We first consider the energy dynamics of the ASV. In particular, we consider an ASV with on-board battery storage, rendering its speed adjustable in real-time. When the ASV's speed exceeds the ``break-even" value where expended power is equal to the power available from the renewable (solar, in our case) resource, the battery's state of charge (SOC) decreases; when the speed falls below the break-even value, the SOC increases. Specifically, the ASV's SOC evolves according to the following differential equation:
\begin{equation}\label{eqn:prelim:energy-dynamics}
    \dot{b}(t) = P_{\mathrm{in}}(t) - k_{\mathrm{h}} - k_{\mathrm{m}} u(t)^3
\end{equation}
where $u(t)$ represents the speed of the ASV, $b(t)$ represents
    the ASV SOC as a function of time, $P_{\mathrm{in}}(t)$ represents the
    solar power input at time $t$, $k_{\mathrm{h}}$ is a constant representing the hotel load
of the ASV (a constant power draw from onboard electronics and sensors), and
$k_{\mathrm{m}}$ represents a gain that represents the power consumed by the ASV's motor.

The ASV's energy dynamics are subject to the following constraints:
\begin{subequations}\label{eqn:prelim:energy-constraints}
    \begin{align}
        b_{\mathrm{min}}    & \leq b(t) \leq b_{\mathrm{max}} \label{eqn:prelim:raw-socconstraints} \\
        u_{\mathrm{min}}    & \leq u(t) \leq u_{\mathrm{max}}
    \end{align}
\end{subequations}
where $b_{\mathrm{min}}$ and $b_{\mathrm{max}}$ represent the minimum and maximum SOC values, respectively, and $u_{\mathrm{min}}$ and $u_{\mathrm{max}}$ represent the minimum and maximum speed values, respectively. In this work, we consider the parameters shown in Table \ref{tab:energy-params}, which are reflective of the SeaTrac SP-48.

\begin{table}[h]
    \caption{SeaTrac SP-48 Energetic Parameters}
    \begin{center}
        \begin{tabular}{|c|c|c|c|}
            \hline
            \textbf{Variable Name}  & \textbf{Symbol} & \textbf{Value} &
            \textbf{Units}                                                                 \\
            \hline
            Hotel Load              & $k_{\mathrm{h}}$           & 10             & $W$               \\
            Motor Constant          & $k_{\mathrm{m}}$           & 83             & $kg \cdot m^{-1}$ \\
            Minimum State of Charge & $b_{\mathrm{min}}$       & 0              & $Wh$              \\
            Maximum State of Charge & $b_{\mathrm{max}}$       & 6500           & $Wh$              \\
            Minimum Speed           & $u_{\mathrm{min}}$       & 0              & $ m \cdot s^{-1}$ \\
            Maximum Speed           & $u_{\mathrm{max}}$       & 2.315          & $ m \cdot s^{-1}$ \\
            \hline
        \end{tabular}
    \end{center}\label{tab:energy-params}
\end{table}

\begin{figure}
    \centering
    \includegraphics[width = \columnwidth]{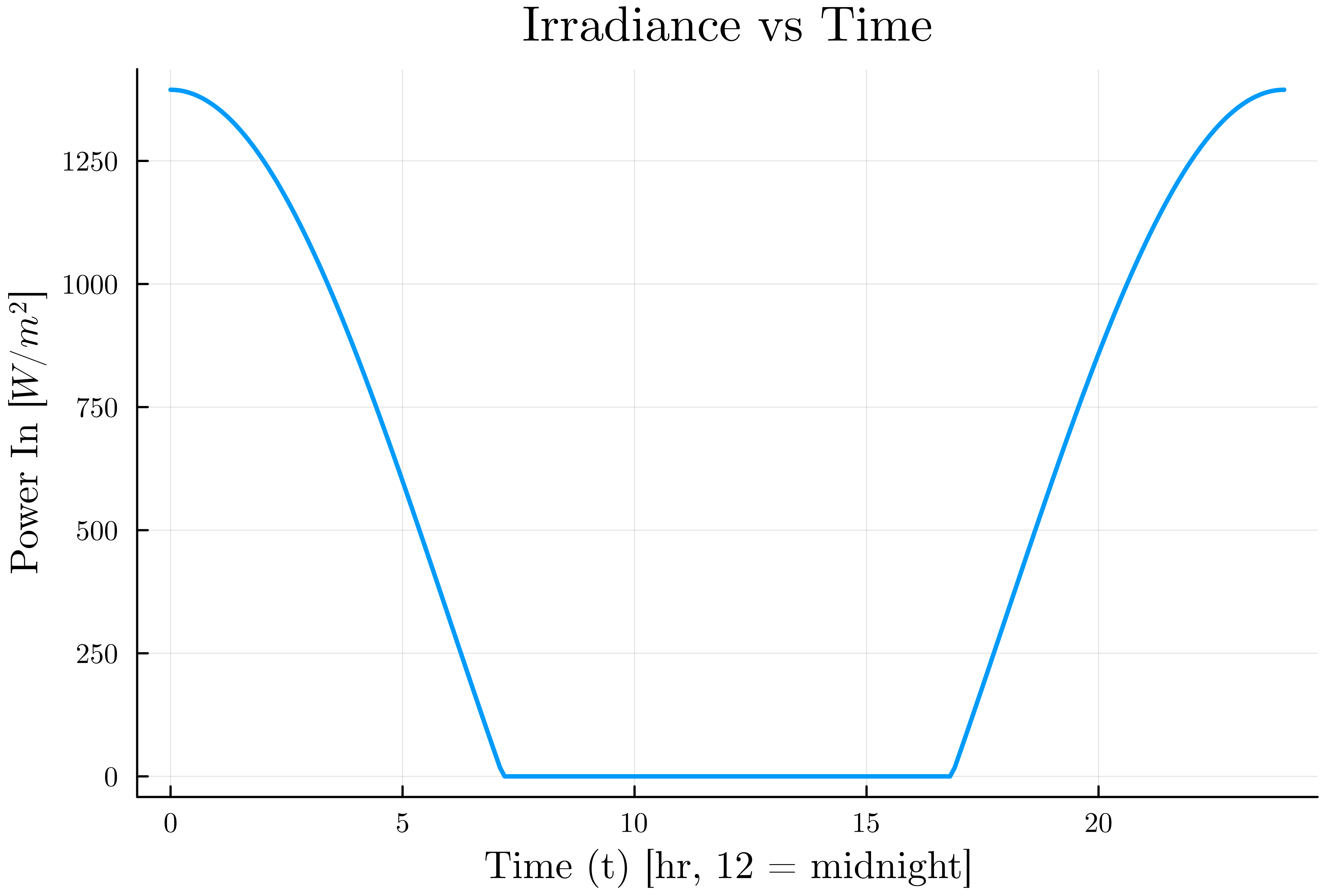}
    \caption{24 Hour Solar Irradiance Model}\label{fig:model-solarirradiance}
\end{figure}

When generating estimates of $P_{\mathrm{in}}(t)$ for the simulations presented in this work, we consider the solar irradiance model described in~\cite{hulstrom2003irradiance}. To summarize, the time-varying solar irradiance, $P_{\mathrm{in}}(t)$, is given by:
\begin{equation}\label{eqn:prelim:irradiance-model}
    P_{\mathrm{in}}(t) = \max \left(0, D_{\mathrm{0}}(t) + D_{\mathrm{1}}(t) \cos\left(\frac{2 \pi t}{T}
    \right) \right)
\end{equation}
where $D_{\mathrm{0}}(t)$ represents the average level of solar irradiance, $D_{\mathrm{1}}(t)$ represents the amplitude of oscillations about the average solar irradiance, and $T$ represents the period with which the solar irradiance model oscillates (in general, $T = 24$ hours). The values of $D_{\mathrm{0}}(t)$ and $D_{\mathrm{1}}(t)$ are determined based upon latitude and time of year. A 24 hour sample of the solar irradiance is shown in Fig.~\ref{fig:model-solarirradiance}.

\subsection{Information (Clarity) Model}\label{sec:methods-stgpkf}
In this work, we formally characterize information through the quantity termed \textbf{\textit{clarity}}, which is a function of position (within the mission domain) and time, is mathematically related to differential entropy~\cite{cover1999elements}, and evolves according to a relatively simple first-order differential equation at every spatial location.

The derivation of the expression for clarity is predicated on the scientific resource of interest evolving according to a Gaussian Process (GP). While this assumption may initially appear highly problematic for several applications -- including our own chosen application, where wind speeds are known not to follow Gaussian distributions -- this complication can be resolved by working with a transformed version of the scientific variable of interest. In fact, we show specifically how this is done for wind speed in Section~\ref{sec:experiment-windspeed}. Because the dynamics in this work are predicated on a GP model for the scientific resource, we begin our presentation with a review of GP essentials, then connect these essential formalisms with clarity.

Within a zero-mean GP (note that the zero mean assumption is made without loss of generality, as a coordinate translation can be used to achieve), a spatiotemporally varying field $ w: \mathbb{R}^{+} \times \mathbb{R}^{2} \rightarrow \mathbb{R} $ follows:
\begin{align}
    w(t,p) &\sim \mathcal{GP}(0, k(t,p,t',p')), \label{eqn:prelim:gp-sample}\\
    k(t,p,t',p') &= \sigma^2 k_{\mathrm{s}}(p,p') k_{\mathrm{t}}(t,t'), \label{eqn:prelim:gp-kernel},
\end{align}
where $t \in \mathbb{R}^{+}$ represents the time, $p \in \mathbb{R}^2$ represents the ASV's position, $ \sigma^2 $ is the process variance, $k_{\mathrm{s}}(p,p')$ represents the spatial kernel, and $k_{\mathrm{t}}(t,t')$ represents the temporal kernel. The spatial and temporal kernels provide parametric models for uncertainty as functions of the spatial and temporal (respectively) separation between measurements and a point of interest $ \sigma^2 $ is the process variance. In this work, we utilize a Matern-1/2 kernel structure \cite{wackernagel2003multivariate}, where:
\begin{align}
    k_{\mathrm{s}}(p,p') &= \exp\left( -\frac{\lVert p - p' \rVert}{\lambda_{\mathrm{s}}} \right) \label{eqn:prelim:gp-kernel_spatial}, \\
    k_{\mathrm{t}}(t,t') &= \exp\left( -\frac{\lVert t - t' \rVert}{\lambda_{\mathrm{t}}} \right) \label{eqn:prelim:gp-kernel_temporal},
\end{align}
where $ \lambda_{\mathrm{s}} $ and $ \lambda_{\mathrm{t}} $ are the spatial and temporal length scales, respectively. In this work, we set $\lambda_{\mathrm{s}} = 0.75 \text{ km}$ and $\lambda_{\mathrm{t}} = 0.25 \text{ hour}$ based on historical wind patterns in Raleigh, North Carolina.

To efficiently characterize the evolution of the GP's prediction mean and variance over time (which ultimately must be performed in real time within the adaptive mission planner), a spatiotemporal Gaussian Process Kalman Filter (STGPKF)~\cite{todescatoEfficientSpatiotemporalGaussian2020} is used. The STGPKF extends the GP Kalman Filter (GPKF) to estimate spatiotemporal fields efficiently by leveraging the equivalence between Gaussian processes (GPs) and stochastic differential equations (SDEs). This approach enables computationally scalable inference while preserving probabilistic uncertainty quantification.

By leveraging the Wiener-Khinchin theorem, we characterize the temporal kernel via a state-space representation~\cite{sarkka2013spatiotemporal}:
\begin{subequations}
    \begin{align}
        \frac{dX_{\mathrm{t}}}{dt} &= A X_{\mathrm{t}} + w_{\mathrm{t}}, \quad w_{\mathrm{t}} \sim \mathcal{N}(0, Q)\label{eqn:prelim:state-space-dynamics}\\
        w(p,t) &= C X_{\mathrm{t}} + v_{\mathrm{t}}, \quad v_{\mathrm{t}} \sim \mathcal{N}(0, R)\label{eqn:prelim:state-space-observation}
    \end{align}
\end{subequations}
where $ X_{\mathrm{t}} \in \mathbb{R}^{N \times 1} $ is the latent state, with $N$ representing the number of spatial locations and each element representing the estimate of the wind field at a specific spatial location, $ A \in \mathbb{R}^{N \times N} $ is the state transition matrix, $ Q \in \mathbb{R}^{N \times N} $ is the process noise covariance, $ C \in \mathbb{R}^{1 \times N} $ is the observation matrix, and $ R \in \mathbb{R} $ is the measurement noise covariance. The state-space model is represented in discrete time as:
\begin{equation}\label{eqn:prelim:discrete-state-space}
    X_{\mathrm{t}+1} = A_{\mathrm{d}} X_{\mathrm{t}} + w_{\mathrm{t}}, \quad w_{\mathrm{t}} \sim \mathcal{N}(0, Q_{\mathrm{d}})
\end{equation}
where $ A_{\mathrm{d}} = e^{A \Delta T} $ and $ Q_{\mathrm{d}} $ is the corresponding discrete noise covariance.

Spatial correlations are captured using the spatial covariance matrix:
\begin{equation}\label{eqn:prelim:spatial-covariance}
    K_{\mathrm{gg}} = [k_{\mathrm{s}}(p_i, p_j)]_{i,j=1}^N
\end{equation}
where $ K_{\mathrm{gg}} $ is factorized as $ K_{\mathrm{gg}} = S S^T $ for computational efficiency. The Kalman filter is then applied to estimate the wind field at all locations, updating as new measurements arrive.

Given the estimates made available via the STGPKF, it is now possible to characterize clarity, along with a set of first-order differential equations that describe the evolution of clarity. In particular, clarity is an information-theoretic measure used to quantify the quality of an estimate. Clarity, denoted as $q(p,t)$, assumes a scalar value in $[0,1]$ at each point in space ($p$) and time ($t$), with higher clarity indicating lower uncertainty~\cite{agrawalSensorBasedPlanningControl2023}. In fact, the value of $q(x,t)$ is related to differential entropy $h(x,t)$ through:
\begin{equation}\label{eqn:prelim:clarity-definition}
    q(p,t) = \left( 1 + \frac{e^{2h(x(p,t))}}{(2\pi e)^n} \right)^{-1},
\end{equation}
\noindent where $x(p,t)$ is the scientific variable of interest at location $p$ and time $t$. For a scientific quantity following a GP model, with a prediction variance of $\Sigma(p,t)$, the value of clarity reduces to:
\begin{equation}\label{eqn:prelim:clarity-gaussian}
    q(p,t) = \frac{1}{1 + \Sigma(p,t)}.
\end{equation}

The aforementioned connection between the STGPKF and clarity gives rise to the following closed-form, first-order differential equation model that describes the evolution of clarity at each location $p$:
\begin{equation}\label{eqn:methods:clarity-dynamics}
    \dot{q}(p,t) = S(p_{\mathrm{r}}(t), p)(1 - q(p,t))^2 - D(p, q(p,t))
\end{equation}
where $S(p_{\mathrm{r}}(t),p)(1-q(p,t))^2$ denotes the information gain from new measurements taken at the ASV position $p_{\mathrm{r}}$ and $D(p,q(p,t))$ represents the natural decay of clarity over time due to environmental change and the growth in uncertainty associated with this. The sensing function follows the following equation:
\begin{equation}\label{eqn:methods:sensing-function}
    S(p_{\mathrm{r}}, p) = \frac{k_{\mathrm{s}}(p_{\mathrm{r}}, p)^2}{k_{\mathrm{s}}(p,p) \left(k_{\mathrm{s}}(p_{\mathrm{r}}, p_{\mathrm{r}}) + \sigma_{\mathrm{m}}^{2}\right) - k_{\mathrm{s}}(p_{\mathrm{r}}, p)^2}
\end{equation}
where $k_{\mathrm{s}}$ is the spatial kernel function and $\sigma_{\mathrm{m}}^{2}$ is the measurement variance (consistent with the sensor noise $R$ defined in Eqn.~\ref{eqn:prelim:state-space-observation}). Clarity decay follows the following equation:
\begin{equation}\label{eqn:methods:decay-function}
    D(p, q) = \frac{2}{\lambda_{\mathrm{t}}} \left( (\sigma^2 + 1)q^2 - q \right)
\end{equation}
where $\lambda_{\mathrm{t}}$ and $\sigma^2$ capture temporal correlation characteristics from the GP kernel in Eqn.~\ref{eqn:prelim:gp-kernel}.

\subsection{Wind Speed Transformation}\label{sec:experiment-windspeed}
As described in Sec.~\ref{sec:methods-stgpkf}, the STGPKF assumes a zero-mean Gaussian distribution for the scientific variable of interest. However, the application considered in this work focuses on characterizing wind speed, which notably will not follow a zero-mean Gaussian distribution.

To transform the collected wind speed measurements to a zero-mean Gaussian distribution, we first make the assumption that the measured wind speed follows a Weibull distribution (as is commonly assumed in wind modeling literature~\cite{seguro2000modern},~\cite{tuller1984characteristics}).

The cumulative distribution function (CDF) of the Weibull distribution is given by:
\begin{equation}\label{eqn:prelim:weibullcdf}
    \Phi_{\mathrm{w}}(x; \lambda, k) = 1 - e^{-\left(\frac{x}{\lambda}\right)^k}
\end{equation}
where $x$ is the wind speed, $\lambda$ is the scale parameter, and $k$ is the shape parameter. For this work, we fix $k = 2.30$ based upon historical wind data available for Jordan Lake from NOAA~\cite{noaa_cdo_findstation}. Treating the forecasted sustained wind speed ($w_{\mathrm{forecast}}$) as the modal wind speed for calibration, we solve for $\lambda$ using
\begin{equation}\label{eqn:prelim:weibulllambda}
    w_{\mathrm{forecast}} = \lambda \left( \frac{k-1}{k} \right)^{\frac{1}{k}}
\end{equation}

The inverse CDF of the Gaussian distribution is given by:
\begin{equation}\label{eqn:prelim:inversegaussian}
    \Phi_{\mathrm{g}}^{-1}(x; \mu, \sigma) = \mu + \sigma \sqrt{2} \text{erf}^{-1}(2x - 1)
\end{equation}
where $\mu$ is the mean and $\sigma$ is the standard deviation of the Gaussian distribution. For this work, we take $\mu = 0$ and $\sigma = 1$. We can then transform the wind speed data to a zero-mean Gaussian distribution by:
\begin{equation}\label{eqn:prelim:windtransform}
    w_{\mathrm{transformed}} = \Phi_{\mathrm{g}}^{-1}\left( \Phi_{\mathrm{w}}(w(p,t); \lambda, k) \right)
\end{equation}
This value $w_{\mathrm{transformed}}$ is then used as the input to the STGPKF.

\section{Adaptive Mission Planning Methodology}\label{sec:methods}
\subsection{Problem Statement}\label{sec:methods-problem}
We consider the problem of exploring a mission domain $\mathcal{D} \subset \mathbb{R}^2$ using a solar-powered autonomous surface vehicle (ASV) tasked with characterizing a spatiotemporally varying wind field. The objective is to allocate sensing effort toward regions where the wind speed $w(p,t)$ is close to a desired rated value $w_{\mathrm{rated}}$, while respecting energetic and operational constraints.

\begin{figure*}[h]
    \centering
    \includegraphics[width=\textwidth]{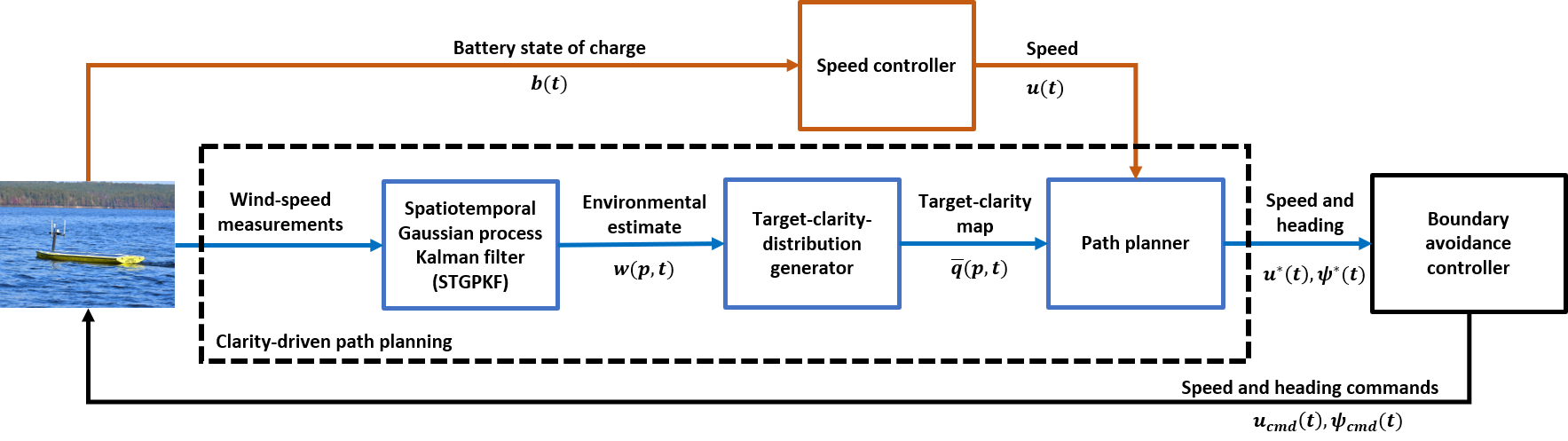}
    \caption{Block diagram of the proposed integrated control framework. Blue blocks correspond to clarity-driven path planning.}\label{fig:blockdiagram}
\end{figure*}

To achieve this, we propose the integrated control framework shown in Fig.~\ref{fig:blockdiagram}. The framework consists of two primary layers:  
\begin{itemize}
    \item \textbf{Energetic control (orange blocks)} ensures that the ASV operates within feasible state-of-charge (SOC) limits and respects mission-duration energy budgets.
    \item \textbf{Clarity-driven path planning (blue blocks)} plans and executes trajectories that maximize information acquisition according to a clarity-driven objective.
\end{itemize}
Together, these components form a resource-aware, information-theoretic exploration architecture.

\subsection{Speed Controller}\label{sec:methods-speedcontrol}
The ASV’s speed controller aims to maximize the distance traveled, while respecting energy (SOC) constraints throughout the mission. For short-duration missions considered in this work (within a single day), we assume speed optimization for persistent endurance is beyond scope (see~\cite{govindarajan2025fusion} for a persistent formulation). Nevertheless, energy constraints must still be enforced.

\begin{figure}[h]
    \centering
    \includegraphics[width = \columnwidth]{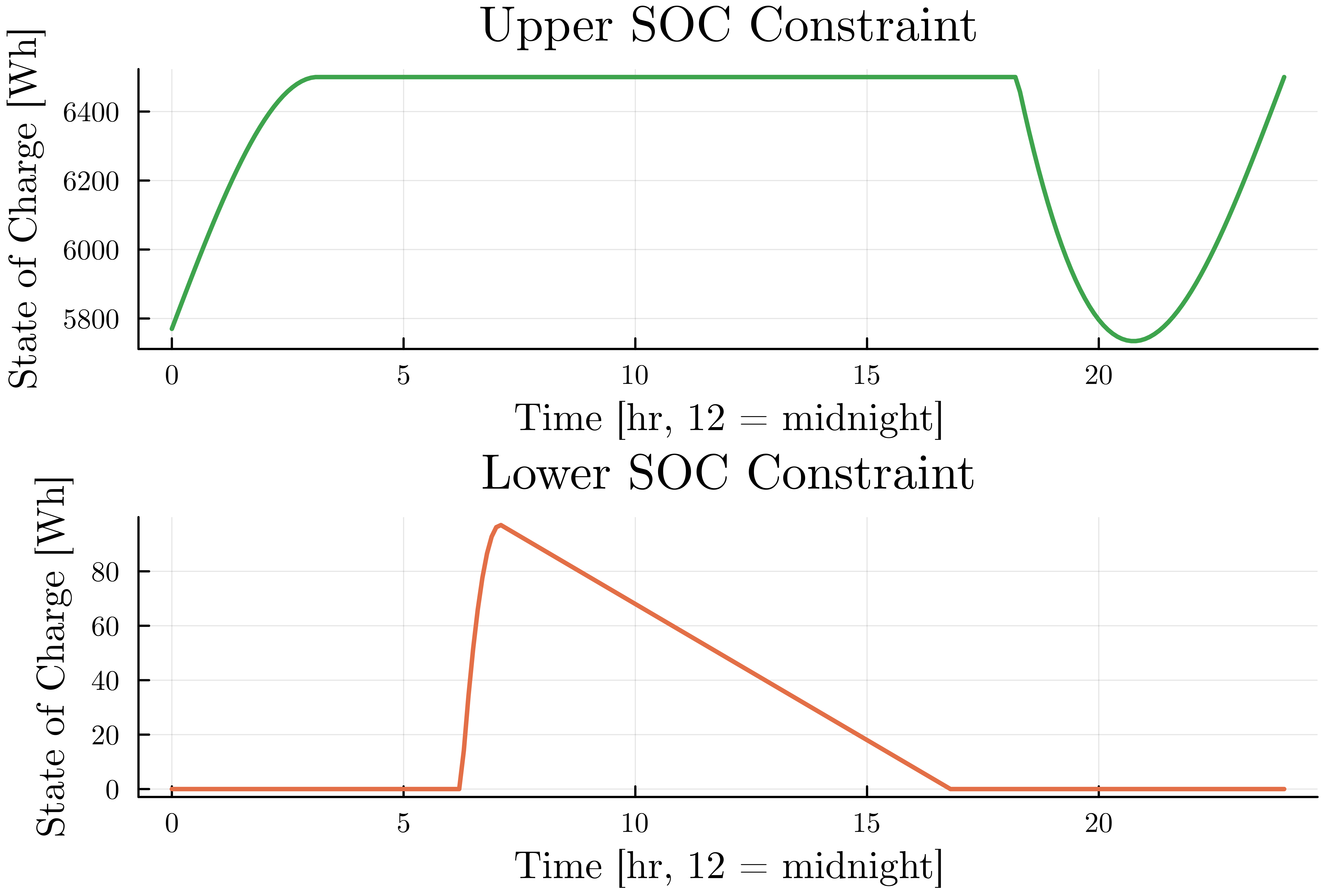}
    \caption{Example of tightened SOC constraints computed over a 24-hour period.}\label{fig:constraintbarriers}
\end{figure}

For the energy dynamics in Eqns.~\ref{eqn:prelim:energy-dynamics}–\ref{eqn:prelim:raw-socconstraints}, pointwise SOC constraint satisfaction does not guarantee future feasibility. To prevent constraint violation, we employ the tightened-constraint formulation from~\cite{govindarajan2025fusion}. The lower and upper constraint functions are computed from the energy deficit and surplus:
\begin{subequations}
    \begin{align}
        \epsilon_{-}(t) &= \int_{t_{\mathrm{0}}}^{t} (k_{\mathrm{h}} - P_{\mathrm{in}}(t)) dt, \label{eqn:methods:energydeficit}\\
        \epsilon_{+}(t) &= \int_{t_{\mathrm{0}}}^{t} (P_{\mathrm{in}}(t) - k_{\mathrm{h}} - k_{\mathrm{m}} u_{\mathrm{max}}^3) dt. \label{eqn:methods:energysurplus}
    \end{align}
\end{subequations}
These quantities are used to compute the tightened constraints $b_{\mathrm{l}}(t)$ and $b_{\mathrm{u}}(t)$ as defined in Eqns.~\ref{eqn:methods:lower_barrier}–\ref{eqn:methods:upper_barrier}:
\begin{align}
    b_{\mathrm{l}}(t_{\mathrm{1}}) &= \begin{cases}
        \sup\limits_{t_{\mathrm{2}} \in [t_{\mathrm{1}}, t_{\mathrm{f}}]} (\epsilon_{-}(t_2) - \epsilon_{-}(t_1)), & \text{if } \sup > 0\\
        0, & \text{otherwise}
    \end{cases}\label{eqn:methods:lower_barrier}\\
    b_{\mathrm{u}}(t_{\mathrm{1}}) &= \begin{cases}
        \sup\limits_{t_{\mathrm{2}} \in [t_{\mathrm{1}}, t_{\mathrm{f}}]} (\epsilon_{+}(t_2) - \epsilon_{+}(t_1)), & \text{if } \sup > 0\\
        0, & \text{otherwise}
    \end{cases}\label{eqn:methods:upper_barrier}
\end{align}
where $t_{\mathrm{1}}$ is the current time at which we are evaluating the tightened constraints, representing the ``starting point'', and $t_{\mathrm{2}} \in [t_{\mathrm{1}}, t_{\mathrm{f}}]$ is a time representing some time in the future. A sample of these tightened constraints is shown in Fig.~\ref{fig:constraintbarriers}. Having derived these tightened constraints, it has been shown in our prior conference paper~\cite{govindarajan2025fusion}, using indirect optimal control methods, that the resulting optimal control law for maximizing distance traveled takes the form:
\begin{equation}\label{eqn:methods:optimalspeed}
    u(t) =
    \begin{cases}
        u_{\mathrm{max}}, & b(t) \ge b_{\mathrm{u}}(t) \\
        u_{\mathrm{min}}, & b(t) \le b_{\mathrm{l}}(t) \\
        \text{constant}, & \text{otherwise.}
    \end{cases}
\end{equation}
\noindent Thus, while indirect methods lead to a switching control law wherein the optimal speed is constant whenever $b(t)$ is not at its upper or lower limits, indirect methods alone do not provide the \textit{\textbf{value}} of this optimal constant speed.

For short-duration missions, where a constant $u(t)$ will not result in $b(t)$ reaching its saturation limits, $u(t)$ can be specified in one of two ways: (i) as a fixed open-loop speed computed to achieve a target SOC at the end of the mission duration, or (ii) as a time-varying target SOC profile that is tracked via a closed-loop PID controller. In the first approach, the difference between $b(t_{\mathrm{i}})$ and $b(t_{\mathrm{f}})$ (where $t_{\mathrm{i}}$ and $t_{\mathrm{f}}$ are the initial and final times) is used to calculate an average power gain/loss according to:
\begin{equation}
P_{\mathrm{gain}} = \frac{b(t_{\mathrm{f}})-b(t_{\mathrm{i}})}{t_{\mathrm{f}}-t_{\mathrm{i}}}.
\end{equation}
\noindent Similarly, the forecast solar power is averaged over the time window according to:
\begin{equation}
P_{\mathrm{in,ave}} = \frac{1}{t_{\mathrm{f}}-t_{\mathrm{i}}} \int_{t_{\mathrm{i}}}^{t_{\mathrm{f}}} P_{\mathrm{in}}(t) dt.
\end{equation}
\noindent Finally, the required constant speed is calculated by substituting $P_{\mathrm{gain}}$ for $\dot{b}$ and substituting $P_{\mathrm{in,ave}}$ for $P_{\mathrm{in}}$ in the system dynamics, then solving for $u(t)$ as:
\begin{equation}
    u(t) = \big( \frac{P_{\mathrm{in,ave}}-P_{\mathrm{gain}}-k_{\mathrm{h}}}{k_{\mathrm{m}}} \big)^{1/3}.
\end{equation}
\noindent In the second approach, $u(t)$ is calculated as above but is then substituted into the system dynamics to calculate a corresponding state of charge profile, which is treated as a setpoint and termed $b_{\mathrm{des}}(t)$. The setpoint is tracked according to a PID controller:
\begin{eqnarray}
\nonumber u(t) &=& K_{\mathrm{p}} (b_{\mathrm{des}}(t) - b(t)) + K_{\mathrm{i}} \int_{t_{\mathrm{i}}}^{t} (b_{\mathrm{des}}(\bar{t}) - b(\bar{t}))\,d\bar{t} \\
&& + K_{\mathrm{d}} \frac{d}{dt} (b_{\mathrm{des}}(t) - b(t)).
\end{eqnarray}
This second approach allows for robust satisfaction of a terminal state of charge target in the absence of a perfect forecast but requires an additional feedback controller to adjust the ASV's speed in response to solar irradiance that does not match the forecast. Both approaches were implemented and tested in our experiments. 

For long-duration missions, a constant speed will generally cause $b(t)$ to hit its upper and lower limits, thereby requiring speed variation. Furthermore, while the results of~\cite{govindarajan2025fusion} guarantee the optimal speed to be constant \textit{\textbf{within each time interval where $b(t)$ lies strictly between $b_{\mathrm{u}}(t)$ and $b_{\mathrm{l}}(t)$}}, the optimal value of $u(t)$ can differ from one such interval to the next. In~\cite{govindarajan2025fusion}, we have developed a learning-based approach to compute this constant speed through an update at the end of each day, though this is beyond the scope of the current work.

\subsection{Clarity-Driven Path Planning}
Ultimately, the clarity-driven path planning selects an ASV trajectory that aims to minimize the gap between the targeted and actual values of clarity, averaged over the domain. This process is broken into three stages: (i) the spatiotemporal Gaussian Process Filter (STGPKF), (ii) target clarity distribution generator, and (iii) path planner, each of which are detailed here.

\subsubsection{Spatiotemporal Gaussian Process Filter (STGPKF)}\label{sec:methods-clarity}
The Spatiotemporal Gaussian Process Filter (STGPKF) provides a computationally efficient means of estimating and updating a time-evolving clarity map, $q(t,p)$, as new measurements are collected. As discussed in Section~\ref{sec:methods-stgpkf}, the STGPKF leverages the structure of a Gaussian process prior to encode spatial correlations across locations $p \in \mathcal{P}$, while using a recursive Kalman filter formulation to handle temporal evolution and sequential data assimilation.

At each time step $t_{\mathrm{k}}$, the filter propagates the clarity state using the process model in Eqn.~\ref{eqn:prelim:state-space-dynamics}. When a new observation $z_{\mathrm{k}}$ is obtained at a subset of spatial locations, the predicted clarity map is updated using the measurement model in Eqn.~\ref{eqn:prelim:state-space-observation}. Overall, the STGPKF can be viewed as a spatiotemporally extended Kalman filter that embeds the covariance structure of a Gaussian process. This structure enables the clarity field $q(t,p)$ to evolve smoothly in time while remaining spatially correlated across the environment, effectively combining the expressiveness of GPs with the sequential efficiency of Kalman filtering.

\subsubsection{Target Clarity Generation}\label{sec:methods-targetclarity}
To focus exploration toward informative regions, we define a \textbf{\textit{target clarity map}} $\overline{q}(p)$ derived from STGPKF wind estimates. Given the estimated mean wind $\mu(p)$ and a desired rated wind $w_{\mathrm{rated}}$, we define:
\begin{subequations}\label{eqn:methods:targetclaritycalc}
    \begin{align}
        \delta(p,t) &= - \lambda_{\mathrm{cd}} \left( \mu(p,t) - w_{\mathrm{rated}}(p) \right)^2 \label{eqn:methods:targetclarity-delta}\\
        \overline{q}(p,t) &= q_{\mathrm{rated}} e^{\delta(p,t)} \label{eqn:methods:targetclarity-qtarget}
    \end{align}
\end{subequations}
where $\lambda_{\mathrm{cd}}$ controls how rapidly the target clarity decays with distance from $w_{\mathrm{rated}}$. $q_{\mathrm{rated}}$ is the desired clarity level achieved in locations where the wind speed equals the rated wind speed. The resulting $\overline{q}(p,t)$ defines the clarity level the planner seeks to achieve across the domain.

\subsubsection{Clarity-Driven Path Planner}\label{sec:methods-pathplanner}
The path planner computes continuous trajectories that guide the ASV toward regions of high target clarity while maintaining coverage across the mission domain $\mathcal{D} = [0, L_x] \times [0, L_y] \subset \mathbb{R}^2$. To achieve this, we formulate the path planning problem using \textit{ergodic control}, which guarantees that the time-averaged spatial occupancy of the vehicle trajectory matches a desired target spatial density distribution.

\paragraph{Target Spatial Distribution Normalization}
Using the current clarity estimate $q(p,t)$ from the STGPKF and the target clarity map $\overline{q}(p,t)$ from Sec.~\ref{sec:methods-targetclarity}, we compute the local clarity deficit:
\begin{equation}\label{eqn:methods:clarity-deficit}
    \Delta q(p,t) = \max\left(0, \, \overline{q}(p,t) - q(p,t)\right).
\end{equation}
To serve as a target spatial distribution for ergodic trajectory optimization, $\Delta q(p,t)$ must be normalized into a valid spatial probability density function $\Phi(p,t)$ satisfying $\int_{\mathcal{D}} \Phi(p,t) \, dp = 1$:
\begin{equation}\label{eqn:methods:normalized-tsd}
    \Phi(p,t) = \frac{\Delta q(p,t)}{\int_{\mathcal{D}} \Delta q(p,t) \, dp}.
\end{equation}
Intuitively, $\Phi(p,t)$ defines the desired relative dwell-time density of the vehicle across the domain; regions where clarity lags significantly behind the target receive proportionally higher spatial probability weighting.

\paragraph{Trajectory Statistics and Fourier Basis Projection}
Let $p(\tau) \in \mathcal{D}$ denote the continuous position trajectory of the ASV for $\tau \in [0, t]$. The empirical spatial density $C(p,t)$ generated by the vehicle trajectory up to time $t$ is represented using the Dirac delta distribution:
\begin{equation}\label{eqn:methods:empirical-density}
    C(p,t) = \frac{1}{t} \int_{0}^{t} \delta\left(p - p(\tau)\right) d\tau.
\end{equation}

To compare the empirical trajectory distribution $C(p,t)$ with the target distribution $\Phi(p,t)$, both spatial functions are projected onto a finite set of orthonormal Fourier cosine basis functions $\{f_k(p)\}_{k \in \mathcal{K}}$, where $k = (k_1, k_2) \in \mathcal{K}$ indexes discrete 2D spatial frequency modes:
\begin{equation}\label{eqn:methods:cosine-basis}
    f_k(p) = \frac{1}{h_k} \cos\left(\frac{k_1 \pi p_x}{L_x}\right) \cos\left(\frac{k_2 \pi p_y}{L_y}\right),
\end{equation}
with $h_k$ serving as the spatial normalization constant ensuring $\int_{\mathcal{D}} f_k(p)^2 dp = 1$:
\begin{equation}\label{eqn:methods:cosine-normalization}
    h_k = \begin{cases} 
        \sqrt{L_x L_y}, & k_1 = 0, \, k_2 = 0, \\ 
        \sqrt{\frac{L_x L_y}{2}}, & (k_1 = 0, \, k_2 > 0) \quad \text{or} \quad (k_1 > 0, \, k_2 = 0), \\ 
        \sqrt{\frac{L_x L_y}{4}}, & k_1 > 0, \, k_2 > 0.
    \end{cases}
\end{equation}

The corresponding Fourier spectral coefficients $c_k(t)$ (trajectory coverage statistics) and $\phi_k(t)$ (target spatial statistics) are computed via inner products over $\mathcal{D}$:
\begin{align}
    c_k(t) &= \int_{\mathcal{D}} C(p,t) f_k(p) \, dp = \frac{1}{t} \int_{0}^{t} f_k\left(p(\tau)\right) d\tau, \label{eqn:methods:fourier_ck} \\
    \phi_k(t) &= \int_{\mathcal{D}} \Phi(p,t) f_k(p) \, dp. \label{eqn:methods:fourier_phik}
\end{align}

\paragraph{Sobolev Ergodic Metric}
The discrepancy between the spatial distribution of the trajectory $c_k(t)$ and the target distribution $\phi_k(t)$ across all spatial frequencies is quantified using the Sobolev space ergodic metric $\mathcal{E}(t)$:
\begin{equation}\label{eqn:methods:ergodic-metric}
    \mathcal{E}(t) = \sum_{k \in \mathcal{K}} \Lambda_k \left( c_k(t) - \phi_k(t) \right)^2,
\end{equation}
where $\Lambda_k$ is a Sobolev weighting factor that places higher penalty on low-frequency spatial mode mismatches:
\begin{equation}\label{eqn:methods:frequency-weighting}
    \Lambda_k = \left( 1 + k_1^2 + k_2^2 \right)^{-\frac{3}{2}}.
\end{equation}
Because $\Lambda_k$ decays rapidly with higher spatial frequencies $\|k\|$, minimizing $\mathcal{E}(t)$ prioritizes global domain coverage first before resolving fine local density variations.

\paragraph{Spectral Gradient Feedback Control Law}
The ASV is modeled using single-integrator kinematic dynamics, $\dot{p}(t) = \mathbf{u}(t)$, where $\mathbf{u}(t) \in \mathbb{R}^2$ is the velocity vector. To minimize the ergodic metric $\mathcal{E}(t)$ in real time, the path planner computes velocity commands along the negative analytical spatial gradient of the ergodic cost surface:
\begin{equation}\label{eqn:methods:pto-control}
    \mathbf{u}^*(t) = -u^*(t) \frac{\nabla_p \mathcal{E}(p(t))}{\|\nabla_p \mathcal{E}(p(t))\|_2},
\end{equation}
where $u^*(t)$ is the desired vehicle speed provided by the energetic speed controller (Sec.~\ref{sec:methods-speedcontrol}), and the spatial gradient vector $\nabla_p \mathcal{E}(p(t)) = \left[ \frac{\partial \mathcal{E}}{\partial p_x}, \frac{\partial \mathcal{E}}{\partial p_y} \right]^\top$ is evaluated analytically as:
\begin{equation}\label{eqn:methods:ergodic-gradient}
    \nabla_p \mathcal{E}(p(t)) = 2 \sum_{k \in \mathcal{K}} \Lambda_k \left( c_k(t) - \phi_k(t) \right) \nabla_p f_k(p(t)).
\end{equation}
The basis function spatial gradient $\nabla_p f_k(p(t))$ evaluates to:
\begin{equation}\label{eqn:methods:cosine-gradient}
    \nabla_p f_k(p(t)) = \frac{1}{h_k} \begin{bmatrix} 
        -\frac{k_1 \pi}{L_x} \sin\left(\frac{k_1 \pi p_x(t)}{L_x}\right) \cos\left(\frac{k_2 \pi p_y(t)}{L_y}\right) \\[6pt]
        -\frac{k_2 \pi}{L_y} \cos\left(\frac{k_1 \pi p_x(t)}{L_x}\right) \sin\left(\frac{k_2 \pi p_y(t)}{L_y}\right)
    \end{bmatrix}.
\end{equation}

Finally, the heading angle $\psi^*(t)$ output by the path planner is extracted directly from the velocity components:
\begin{equation}\label{eqn:methods:planner-heading}
    \psi^*(t) = \tan^{-1}\!\left(\frac{u_y^*(t)}{u_x^*(t)}\right).
\end{equation}
Equations~\eqref{eqn:methods:pto-control} and \eqref{eqn:methods:planner-heading} provide continuous feedback control driving the ASV along trajectories that dynamically minimize spatial Fourier mode mismatch relative to the evolving target clarity field.

\subsection{Boundary Avoidance}\label{sec:methods-boundaryavoidance}
Finally, the boundary correction ensures that the ASV remains safely within the convex mission domain $\mathcal{D}$. Given the ASV's position $p_{\mathrm{ASV}}$, the nearest boundary point is:
\begin{equation}\label{eqn:methods:nearest-boundary}
    p_{\mathrm{closest}} = \arg\min_{p \in \mathcal{P}} \lVert p - p_{\mathrm{ASV}} \rVert,
\end{equation}
where $\mathcal{P}$ denotes the domain boundary. Let $c$ be the domain centroid and $d = \lVert p_{\mathrm{closest}} - p_{\mathrm{ASV}} \rVert$ the distance to the boundary. The inward-pointing direction from $p_{\mathrm{closest}}$ toward $c$ is:
\begin{equation}\label{eqn:methods:normal-vector}
    \mathbf{n} = \frac{c - p_{\mathrm{closest}}}{\lVert c - p_{\mathrm{closest}} \rVert}.
\end{equation}

As the ASV approaches the boundary, the nominal velocity $\mathbf{u}^*(t) = u^*(t) \begin{bmatrix}
        \cos\psi^*(t) \\[3pt]
        \sin\psi^*(t)
\end{bmatrix}$ is blended with the inward normal direction to maintain a safe standoff distance:
\begin{equation}\label{eqn:methods:boundary-correction}
    \mathbf{u}_{\mathrm{cmd}}(t)
    =
    \left(1 - \frac{d}{d_{\mathrm{min}}}\right) \mathbf{u}^*(t)
    +
    \left(\frac{d}{d_{\mathrm{min}}}\right)
    \lVert \mathbf{u}^*(t) \rVert \mathbf{n},
\end{equation}
where $d_{\mathrm{min}}$ is the minimum safe standoff distance. When $d \ge d_{\mathrm{min}}$, $\mathbf{u}_{\mathrm{cmd}}(t) = \mathbf{u}^*(t)$. As $d$ decreases below $d_{\mathrm{min}}$, the correction term rotates the velocity vector toward the domain interior while preserving speed magnitude.

Finally, the corrected velocity is converted into corrected speed and heading commands:
\begin{align}
    u_{\mathrm{cmd}}(t) &= \lVert \mathbf{u}_{\mathrm{cmd}}(t) \rVert, \\
    \psi_{\mathrm{cmd}}(t)
    &= \tan^{-1}\!\left(\frac{u_{\mathrm{cmd},y}(t)}{u_{\mathrm{cmd},x}(t)}\right)
\end{align}
These commanded values $(u_{\mathrm{cmd}}(t), \psi_{\mathrm{cmd}}(t))$ are sent to the ASV's low-level speed and heading controllers.

\section{Simulation Study}\label{sec:simulation}
In this section, we use simulation studies to examine the effects of target adaptation and the low-level path planner on sampling behavior, field reconstruction, and clarity deficit. We first isolate these effects in two diagnostic environments and then evaluate their sensitivity to the spatiotemporal correlation scales of the wind field through a Monte Carlo study.

\subsection{Ablation and Sensitivity Study Design}

\subsubsection{Compared Strategies}

We compare the seven strategies summarized in Table~\ref{tab:control_algorithms}. The nonadaptive strategies plan using an initially uniform target clarity distribution, while the adaptive strategies update the target distribution online using the STGPKF estimate. The ground-truth-target strategies replace the estimated target distribution with the target clarity computed directly from the simulated wind field. These strategies provide diagnostic references that isolate errors caused by estimating the target distribution.

For low-level planning, we compare the ergodic controller developed in this work with a discrete receding-horizon branch-and-bound informative path planner (BB-IPP)~\cite{binney2012branch}. BB-IPP uses a five-step planning horizon and seven motion primitives and is re-planned at each controller update. Both planners use the same target clarity distribution for a given strategy.

\begin{table}[t]
    \centering
    \caption{Strategies used in the simulation study.}
    \label{tab:control_algorithms}
    \footnotesize
    \begin{tabular}{p{0.27\columnwidth} p{0.27\columnwidth} p{0.33\columnwidth}}
        \toprule
        \textbf{Strategy} & \textbf{Target distribution} & \textbf{Path planner} \\
        \midrule
        Transect & Uniform & Fixed lawnmower path \\
        Ergodic, nonadaptive & Uniform & Ergodic controller \\
        Ergodic, adaptive & STGPKF estimate & Ergodic controller \\
        BB-IPP, nonadaptive & Uniform & Branch-and-bound IPP \\
        BB-IPP, adaptive & STGPKF estimate & Branch-and-bound IPP \\
        Ergodic, ground truth & Ground-truth field & Ergodic controller \\
        BB-IPP, ground truth & Ground-truth field & Branch-and-bound IPP \\
        \bottomrule
    \end{tabular}
\end{table}

\subsubsection{Simulation Environments}

\begin{figure}[t]
    \centering
    \includegraphics[width=\columnwidth]{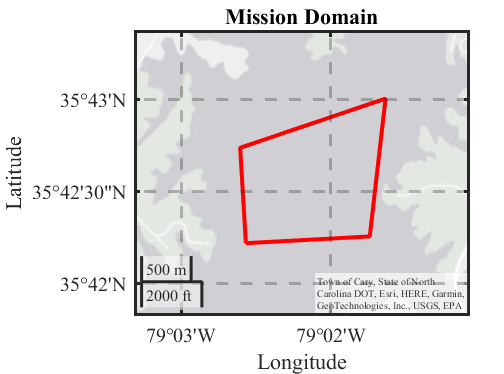}
    \caption{Mission domain at Jordan Lake, North Carolina.}
    \label{fig:jordanlakedomain}
\end{figure}

\begin{figure}[t]
    \centering
    \includegraphics[width=\columnwidth]{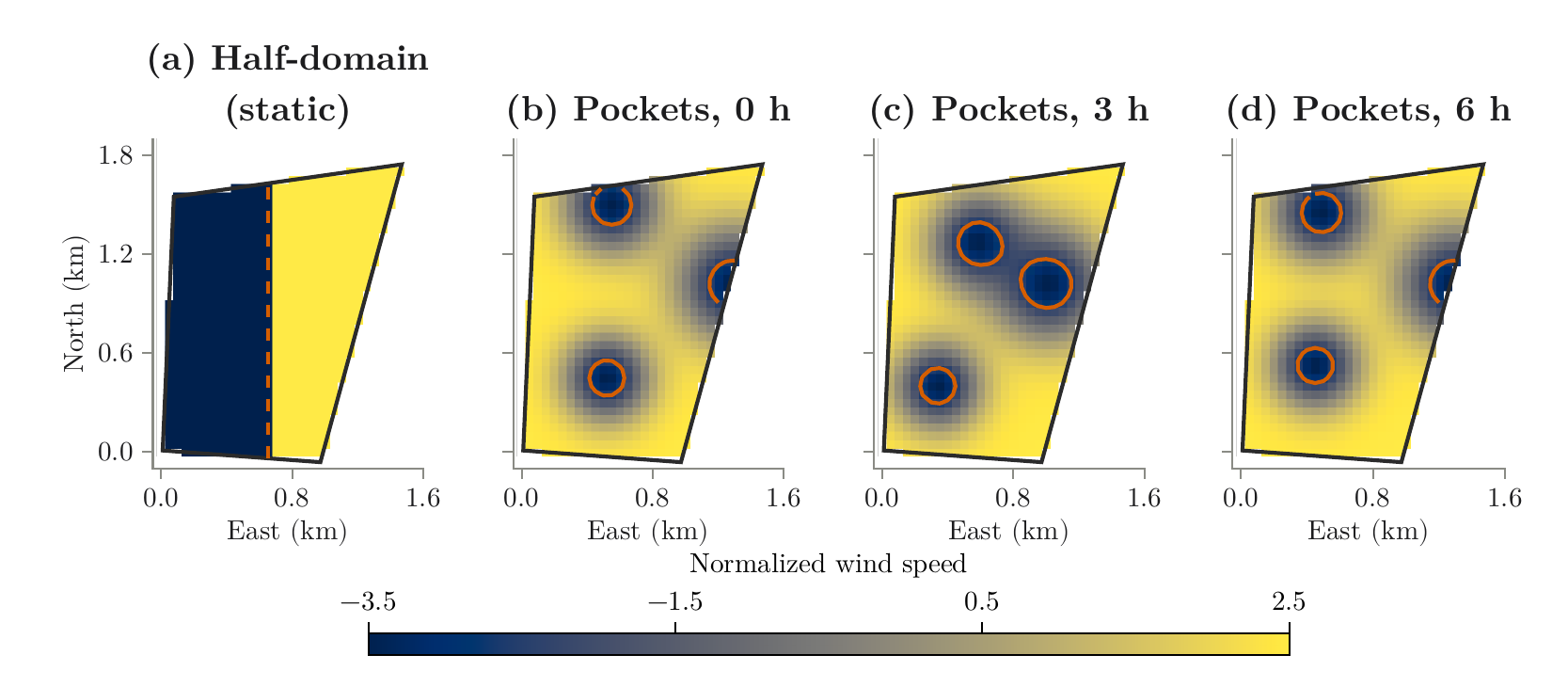}
    \caption{Diagnostic simulation environments. The half-domain case separates favorable and unfavorable wind conditions, while the moving-pocket case contains multiple time-varying regions of near-rated wind.}
    \label{fig:simulation-environments}
\end{figure}

The simulations use the bounded Jordan Lake domain shown in Fig.~\ref{fig:jordanlakedomain}. We consider the two environments shown in Fig.~\ref{fig:simulation-environments}. The half-domain environment is temporally static: approximately half of the navigable domain contains wind near the rated turbine speed, while the remaining half contains strongly off-target wind. This provides a direct test of whether and to what extent the target adaptation moves the vehicle toward the relevant portion of the domain.

The moving-pocket environment contains spatially localized, time-varying pockets of near-rated wind. This case tests whether the controllers can identify and revisit multiple relevant regions while the field changes over time.

\subsubsection{Evaluation Metrics and Parameters}

Performance is evaluated using three complementary metrics. The in-target measurement fraction is the percentage of measurements collected within a prescribed neighborhood of the rated wind speed. Wind reconstruction RMSE measures the accuracy of the estimated wind field over the domain. Ground-truth-target clarity deficit compares the achieved clarity distribution with the target clarity computed from the simulated wind field. Lower RMSE and lower clarity deficit are better.

These metrics capture distinct objectives. The in-target fraction measures sampling concentration, reconstruction RMSE measures global field estimation, and clarity deficit measures how well the trajectory distributes sensing effort according to the target distribution.

Unless otherwise stated, all strategies use the same mission duration, vehicle model, measurement model, STGPKF, and initial condition. The target-selectivity coefficient $\lambda_{\mathrm{cd}}$ controls how quickly target clarity decreases as the estimated wind speed moves away from the rated value. Setting $\lambda_{\mathrm{cd}}=0$ produces a uniform target distribution and recovers the nonadaptive case. Low-level tracking gains, the ergodic spectral truncation, measurement noise, and maximum target clarity are held fixed to isolate the effects of target adaptation, planner selection, and environmental correlation scales.

\subsection{Target-Adaptation Sensitivity}
\label{sec:simulation-lambda-sensitivity}

\begin{figure}[t]
    \centering
    \includegraphics[width=\columnwidth]{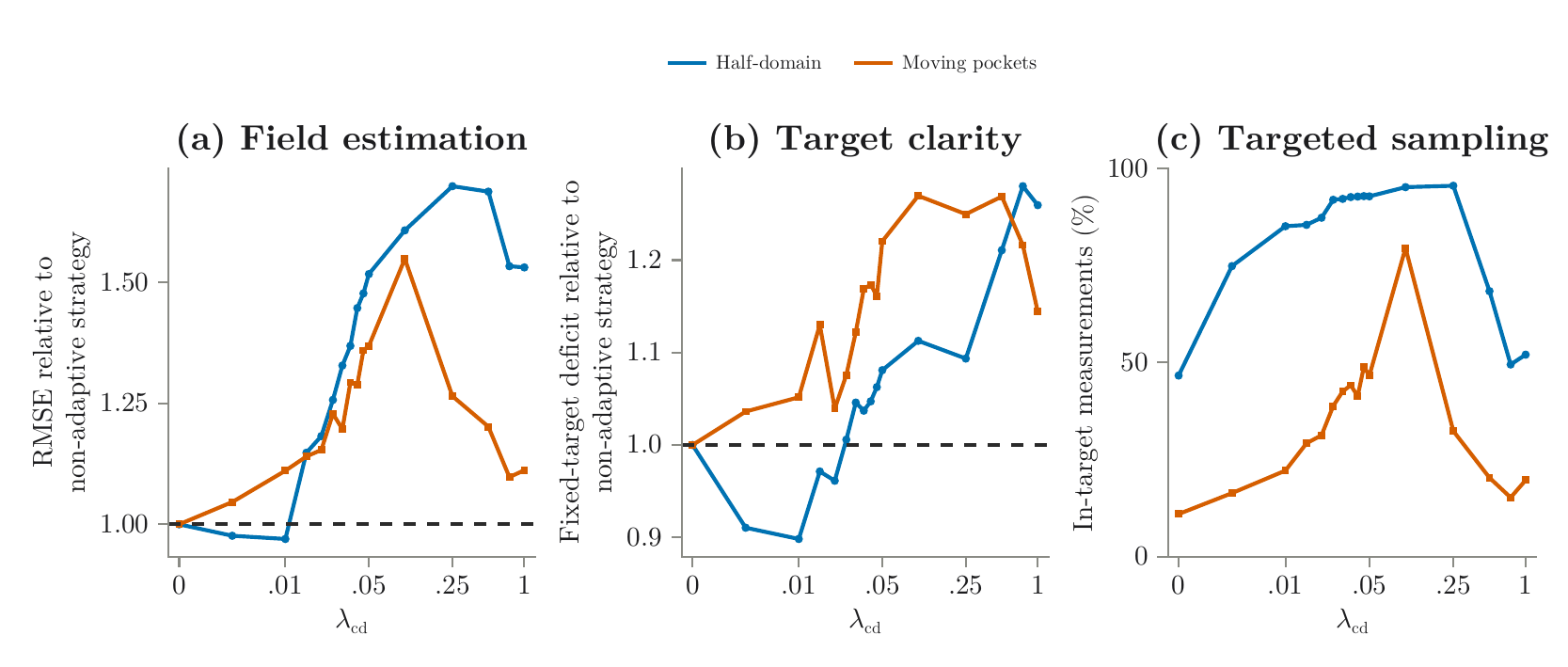}
    \caption{Sensitivity to the target-selectivity coefficient $\lambda_{\mathrm{cd}}$. Moderate selectivity increases sampling near the rated wind speed, while large values can degrade exploration and reconstruction.}
    \label{fig:lambda-cd-sweep}
\end{figure}

Figure~\ref{fig:lambda-cd-sweep} shows the effect of target selectivity. At $\lambda_{\mathrm{cd}}=0$, the adaptive and nonadaptive controllers are identical. Increasing $\lambda_{\mathrm{cd}}$ generally increases the fraction of samples collected near the rated wind speed, but aggressive selectivity can reduce broader exploration and degrade other metrics.

In the half-domain environment, increasing $\lambda_{\mathrm{cd}}$ to $0.01$ increases the ergodic controller's in-target measurement fraction from $46.7\%$ to $85.1\%$. Over the same change, wind reconstruction RMSE decreases from $1.67$ to $1.62$, and ground-truth-target clarity deficit decreases from $0.086$ to $0.077$. This result shows that moderate selectivity can improve both target-directed sampling and clarity deficit when the relevant region is large and static.

In the moving-pocket environment, the tradeoff is stronger. Increasing $\lambda_{\mathrm{cd}}$ to $0.02$ raises the in-target fraction from $11.0\%$ to $31.3\%$, while reconstruction RMSE and ground-truth-target clarity deficit increase by $15.4\%$ and $3.9\%$, respectively. The in-target fraction reaches $79.5\%$ at $\lambda_{\mathrm{cd}}=0.1$ before decreasing to $32.4\%$ at $\lambda_{\mathrm{cd}}=0.25$. The decline at large values indicates that a narrowly concentrated target can cause the controller to overreact to local estimates.

The clarity deficit computed using the controller's estimated target distribution should be interpreted carefully. Because this target is derived from the estimated wind field, regions that have not been adequately explored may be assigned low importance. We therefore use ground-truth-target clarity deficit for the primary comparisons below.

\subsection{Diagnostic Planner Comparisons}

\subsubsection{Half-Domain Environment}

\begin{figure}[t]
    \centering
    \includegraphics[width=\columnwidth]{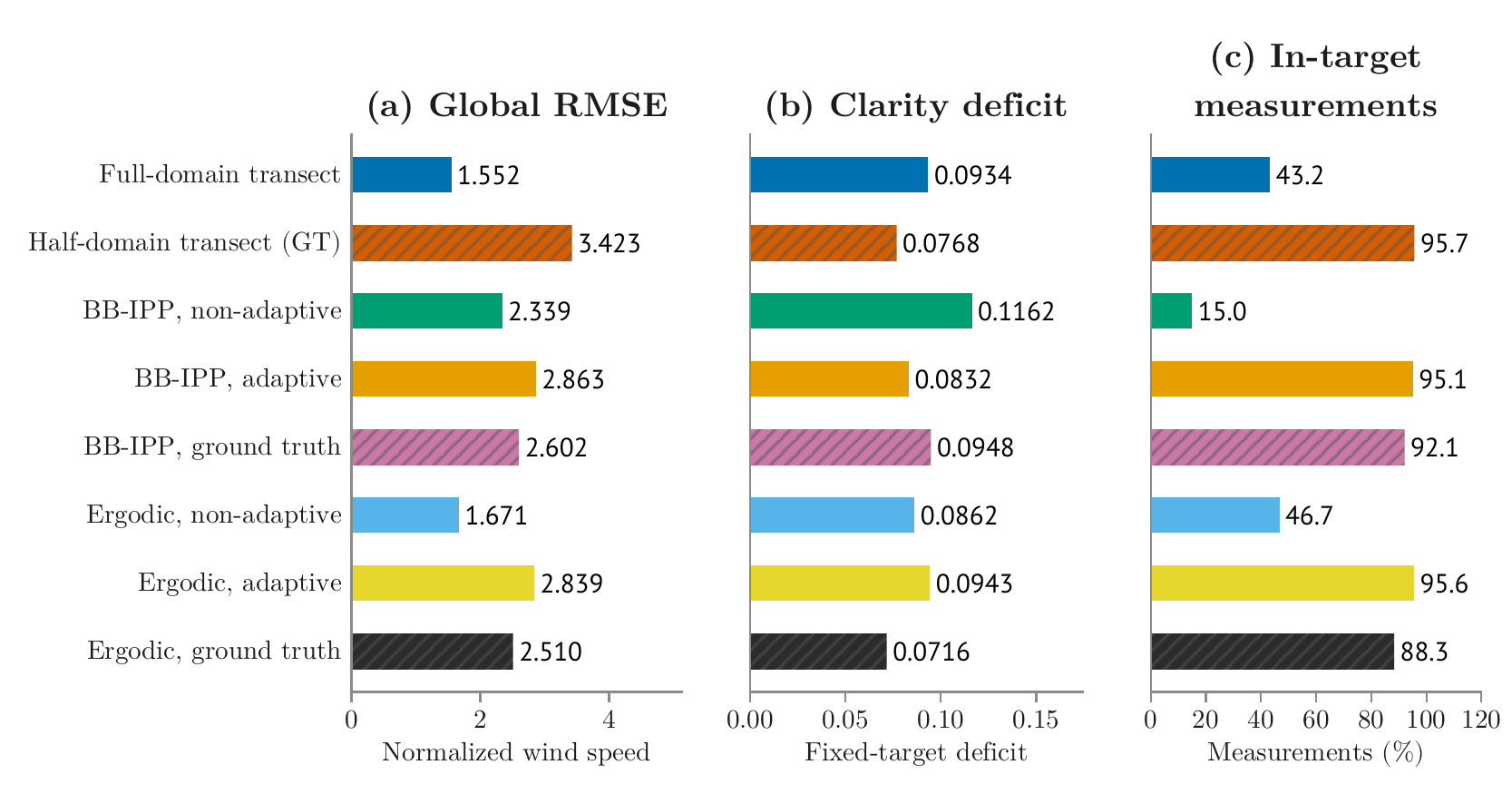}
    \caption{Half-domain results. Restricting the transect to the favorable half and providing the ergodic controller with the ground-truth target both reduce clarity deficit, while online adaptation substantially increases the fraction of near-rated measurements.}
    \label{fig:revision-half-domain}
\end{figure}

Figure~\ref{fig:revision-half-domain} summarizes the half-domain results. As a very basic comparison, we compare the full-domain transect with a transect confined to the favorable half of the domain. Restricting the transect increases the in-target measurement fraction from $43.2\%$ to $95.7\%$ and decreases ground-truth-target clarity deficit from $0.093$ to $0.077$. This confirms that spending more mission time in the relevant half of the domain indeed produces the expected reduction in clarity deficit.

Adaptation produces the same qualitative behavior for both planners. For BB-IPP, the in-target fraction increases from $15.0\%$ under the uniform target to $95.1\%$ with adaptation. For the ergodic controller, it increases from $46.7\%$ to $95.6\%$. The ground-truth-target ergodic strategy achieves the lowest clarity deficit, $0.072$, while the full-domain transect achieves the lowest wind reconstruction RMSE, $1.55$.

The adaptive ergodic controller has a slightly larger ground-truth-target clarity deficit than its uniform counterpart, $0.094$ compared with $0.086$, despite collecting substantially more measurements in the favorable region. The ground-truth-target controller's stronger performance indicates that this discrepancy is caused by error in the estimated target distribution rather than by an inability of the ergodic planner to follow a correct target. Essentially, by targeting specific areas of the domain with near-rated wind conditions, the ergodic adaptive planner does an impressive job of remaining within the target range; however, this is accomplished at the expense of maintaining an accurate estimate elsewhere in the domain, which in turn results in target clarity values that differ from what they ``should be" in those unexplored areas. The restricted-transect diagnostic provides an independent check of this interpretation.

\subsubsection{Moving-Pocket Environment}

\begin{figure}[t]
    \centering
    \includegraphics[width=\columnwidth]{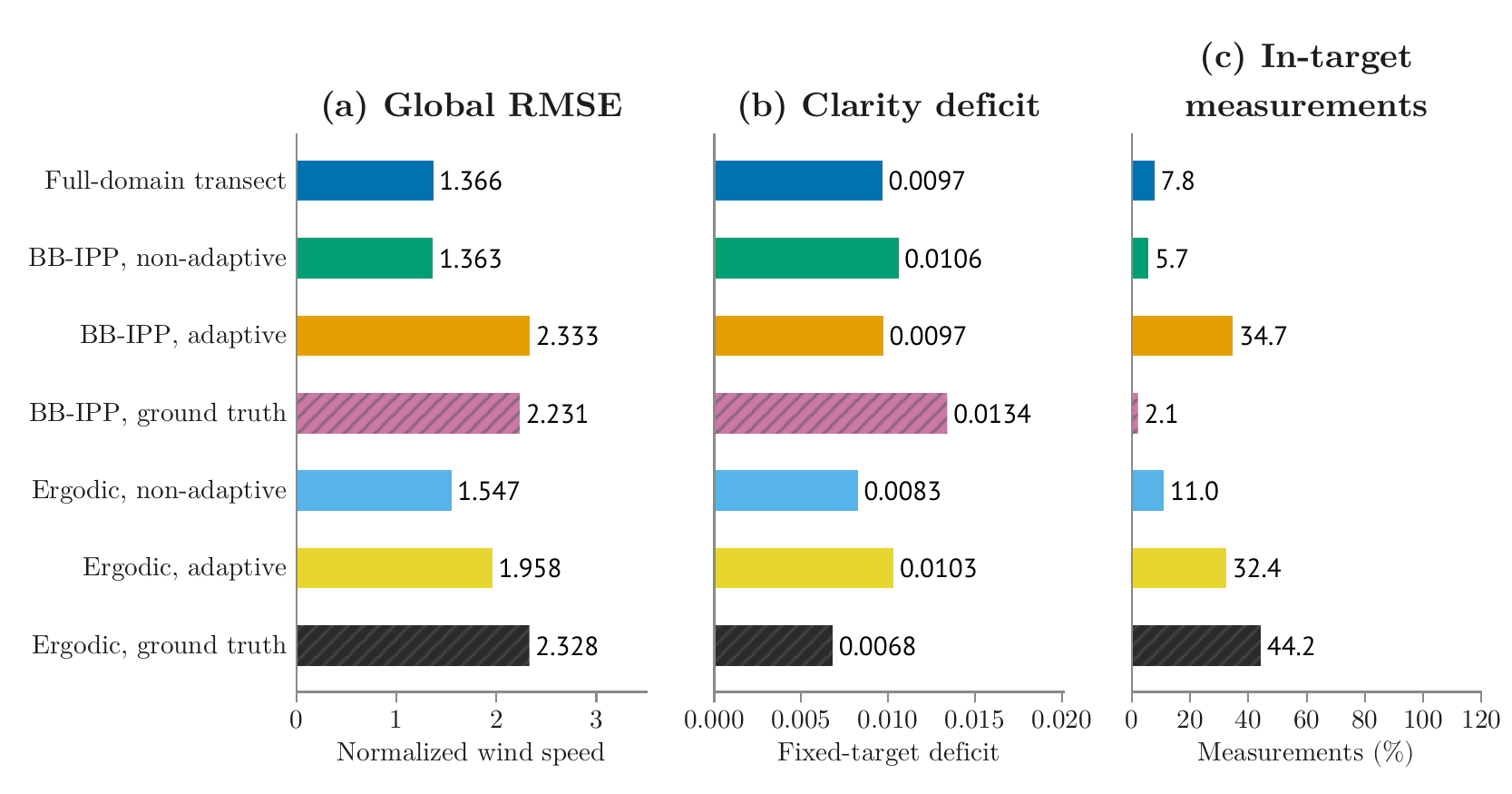}
    \caption{Moving-pocket results. Adaptation increases near-rated sampling for both planners, while the ground-truth-target ergodic controller better distributes sensing effort across multiple pockets.}
    \label{fig:revision-moving-pocket}
\end{figure}

The moving-pocket results are shown in Fig.~\ref{fig:revision-moving-pocket}. Adaptation increases the in-target measurement fraction from $5.7\%$ to $34.7\%$ for BB-IPP and from $11.0\%$ to $32.4\%$ for the ergodic controller. The ground-truth-target ergodic strategy achieves the largest in-target fraction, $44.2\%$, and the lowest clarity deficit, $0.0068$. In comparison, the ground-truth-target BB-IPP strategy collects only $2.1\%$ of its measurements in target.

The trajectories explain this difference. BB-IPP greedily concentrates on a locally valuable pocket and may remain there after the pocket moves or other relevant regions emerge. The ergodic controller distributes sensing effort according to the spatial target distribution and therefore continues to visit multiple pockets. The uniform BB-IPP strategy and the transect achieve the lowest wind reconstruction errors because they maintain broader spatial coverage without concentrating on near-rated conditions.

\subsection{Monte Carlo Sensitivity Study}
\label{sec:simulation-monte-carlo}

\begin{figure*}[t]
    \centering
    \includegraphics[width=\textwidth]{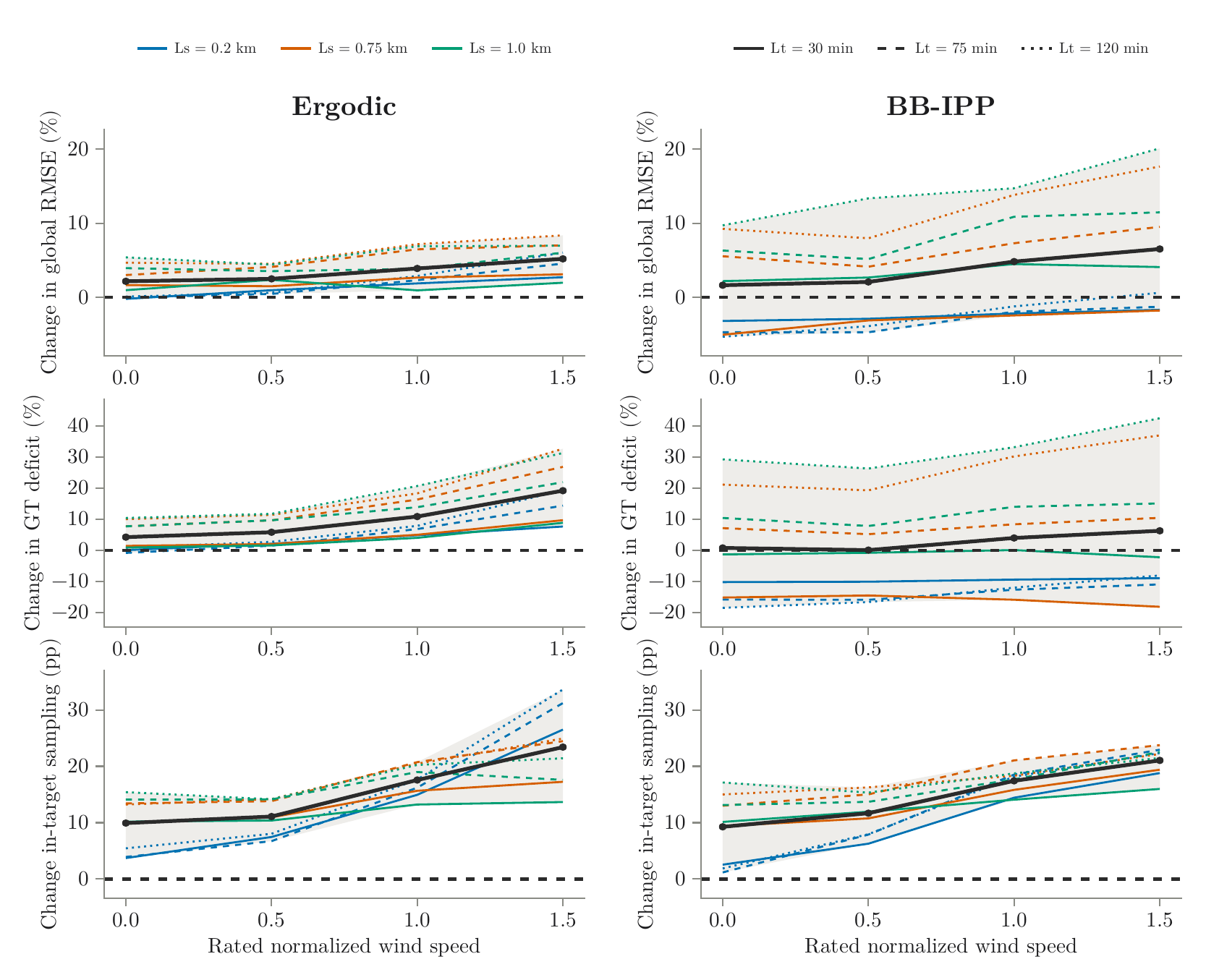}
    \caption{Monte Carlo sensitivity results across rated wind speed, spatial correlation scale $L_s$, and temporal correlation scale $L_t$. Color identifies $W$, line style identifies $L_s$, and marker shape identifies $L_t$.}
    \label{fig:monte-carlo-sensitivity}
\end{figure*}

We next evaluate whether the diagnostic results persist across different wind fields. The Monte Carlo study varies the rated wind speed, spatial correlation scale $L_s$, and temporal correlation scale $L_t$. Each parameter combination is evaluated over multiple random field realizations using identical seeds across strategies. Figure~\ref{fig:monte-carlo-sensitivity} reports the mean performance and variability for each combination.

Across all 36 parameter combinations, adaptation increases the in-target measurement fraction for both planners. For the ergodic controller, the mean increase is $15.50$ percentage points. The associated relative change in wind reconstruction RMSE averages $+3.44\%$, ranging from $-0.21\%$ to $+8.37\%$. For BB-IPP, adaptation increases the in-target fraction by an average of $14.85$ percentage points, while the relative RMSE change averages $+3.77\%$ and ranges from $-5.31\%$ to $+20.09\%$.

The ground-truth-target clarity deficit increases for the adaptive ergodic controller in 35 of the 36 parameter combinations, with a mean relative increase of $10.04\%$. BB-IPP exhibits a more mixed response: adaptation lowers the deficit in 19 combinations and increases it in 17, with a mean relative change of $+2.79\%$. Larger temporal and spatial correlation scales generally reduce reconstruction error because the field varies more smoothly, but they do not change the central result that adaptation consistently shifts measurements toward the target wind conditions.

Using the ground-truth target distribution reduces the clarity deficit relative to the estimated-target adaptive strategies by averages of $8.61\%$ for the ergodic controller and $7.81\%$ for BB-IPP. This improvement is accompanied by average RMSE increases of $5.76\%$ and $5.22\%$, respectively. These results separate the effect of target-estimation error from the behavior of the low-level planner: when provided with the correct target, both planners better match the desired clarity distribution, but this concentration can reduce whole-field reconstruction accuracy.

\subsection{Discussion}
\label{sec:simulation-discussion}

The simulation results support three conclusions. First, online adaptation consistently directs measurements toward the desired wind conditions. This effect appears in both diagnostic environments and across every parameter combination in the Monte Carlo study. Second, target-directed sampling, whole-field reconstruction, and clarity deficit are distinct objectives. Improving one does not necessarily improve the others, particularly when the target distribution is estimated from incomplete measurements. Third, the low-level planner determines how sensing effort is distributed within the target regions. BB-IPP tends to exploit locally valuable regions, while ergodic planning preserves broader coverage of the complete target distribution.

The half-domain simulations address the paradoxical lack of clarity deficit improvement observed with the estimated adaptive target, even when clarity deficit is used explicitly to drive the low-level mission planner. Both the restricted transect and the ground-truth-target ergodic controller reduce the ground-truth clarity deficit as expected. The remaining discrepancy therefore arises from the STGPKF-derived target map, which can underestimate regions that have not yet been sufficiently observed. In an adaptive strategy that intentionally leaves numerous regions unexplored, this can result in significant mismatches between the ground-truth target clarity map (i.e., the ``correct" map) and that which is identified online. This motivates future investigations into definitions of target clarity that are driven by more parameters than just the running prediction mean from the STGPKF.

\section{Experimental Results}\label{sec:experiment}
This section evaluates the framework on the SeaTrac ASV. We consider three mission profiles: 90-minute missions at constant speed, 90-minute missions with closed-loop SOC control, and 3-hour missions with closed-loop SOC control. The first comparison examines targeted sampling without speed adaptation; the latter two examine operation of the planner and energy controller together.

\subsection{Setup}\label{sec:experiment-setup}
Experiments were based on the SeaTrac SP-48 ASV, with the goal of tracking a target wind speed during each experimental run (same as in the simulation studies). An ultrasonic wind sensor was used to measure the wind speed. The ASV was also equipped with GPS for localization and real-time SOC measurements for closed-loop speed control. We tested the ASV at Jordan Lake, North Carolina in the mission domain shown in Fig. \ref{fig:jordanlakedomain}, which is the same mission domain that was used for simulation studies.

For the experiments presented in this section, the forecasted wind speed was obtained from Jordan Lake State Recreation Area weather station data published by NOAA~\cite{noaa_cdo_findstation}. For all of the experimental missions, the forecasted wind speed was 10 mph (4.47 m/s), which is used to calibrate the Weibull scale rather than being assumed to map exactly to zero in Gaussian coordinates. Using Eqn. \ref{eqn:prelim:weibulllambda}, we compute the corresponding Weibull scale parameter to be $\lambda = 5.73$. This value is used in Eqn. \ref{eqn:prelim:windtransform} to transform the wind speed data for input to the STGPKF.

\subsection{90 Minute Missions Without Closed-Loop SOC Control}\label{sec:experiment-no-speed}
We first consider 90-minute missions at constant speed and without closed-loop SOC control. These runs isolate the planner's ability to target wind speeds near the rated value used for wind-turbine siting. The rated wind speed defines the scientific objective; it is unrelated to the solar energy harvested by the ASV.

For these missions, we fix $u(t) = u_{\mathrm{max}}$ for all $t$. This configuration is representative of scenarios where energy constraints can be neglected, such as short-duration missions where battery capacity is sufficient regardless of harvesting efficiency. In such cases, traveling at maximum speed increases the available path length, although this alone does not guarantee greater coverage or information gain.

\begin{figure}
    \centering
    \includegraphics[width=\columnwidth]{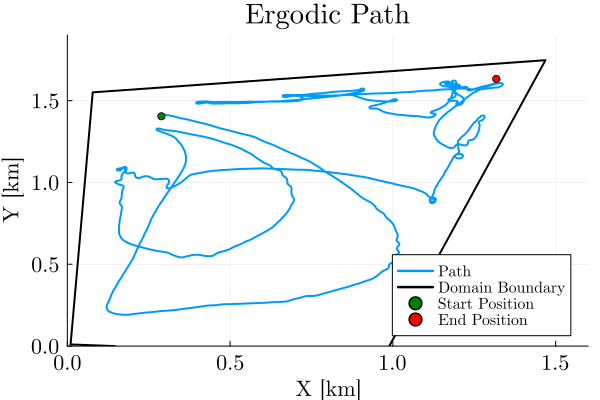}
    \caption{ASV trajectory for the 90-minute ergodic mission without speed control. The trajectory concentrates in the northeast portion of the domain, where measured wind speeds approach the rated value.}\label{fig:90-const-path}
\end{figure}

\begin{figure}
    \centering
    \includegraphics[width=\columnwidth]{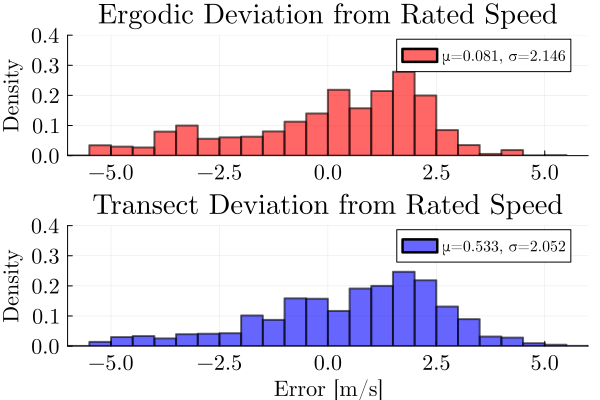}
    \caption{Measured-minus-rated wind-speed distributions for the 90-minute missions without speed control. The mean deviations are $0.081\,\mathrm{m/s}$ for ergodic control and $0.533\,\mathrm{m/s}$ for the transect.}\label{fig:90-const-measurement-error}
\end{figure}

The ergodic trajectory in Fig.~\ref{fig:90-const-path} concentrates in the northeast portion of the domain after the ASV measures wind speeds near the rated value there. The measured-minus-rated distributions in Fig.~\ref{fig:90-const-measurement-error} show a greater concentration near zero for ergodic control than for the transect. These runs therefore demonstrate targeted sampling before speed adaptation is introduced.

\subsection{90 Minute Missions With Closed-Loop SOC Control}\label{sec:experiment-ninety-min}
We next allow the ASV speed to vary according to the SOC controller in Sec.~\ref{sec:methods-speedcontrol}. These missions test whether targeted path planning continues to operate while speed changes to track the desired SOC profile.

For these missions, we compute a nominal SOC trajectory from a user-defined terminal SOC and a constant-speed assumption under the nominal solar resource in Sec.~\ref{sec:methods-speedcontrol}. A PID controller tracks this trajectory in real time by adjusting $u(t)$. The experimental gains are $K_{\mathrm{p}} = 0.01$, $K_{\mathrm{i}} = 0.0001$, and $K_{\mathrm{d}} = 0.001$.

\begin{figure}
    \centering
    \includegraphics[width=\columnwidth]{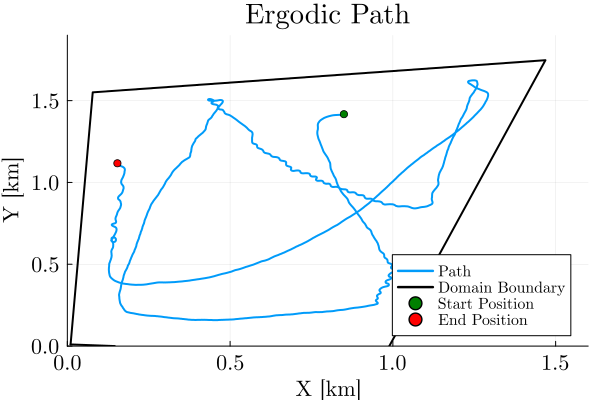}
    \caption{ASV trajectory for the 90-minute ergodic mission with speed control. The trajectory covers a larger portion of the domain while the wind-field estimate develops.}\label{fig:90-variable-path}
\end{figure}

\begin{figure}
    \centering
    \includegraphics[width=\columnwidth]{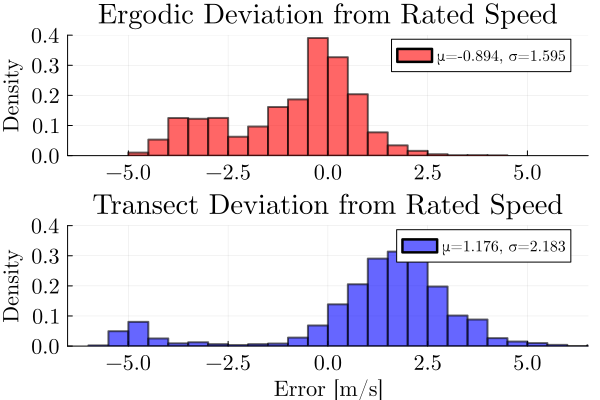}
    \caption{Measured-minus-rated wind-speed distributions for the 90-minute variable-speed missions. The mean deviations are $-0.894\,\mathrm{m/s}$ for ergodic control and $1.176\,\mathrm{m/s}$ for the transect.}\label{fig:90-variable-measurement-error}
\end{figure}

\begin{figure}
    \centering
    \includegraphics[width=\columnwidth]{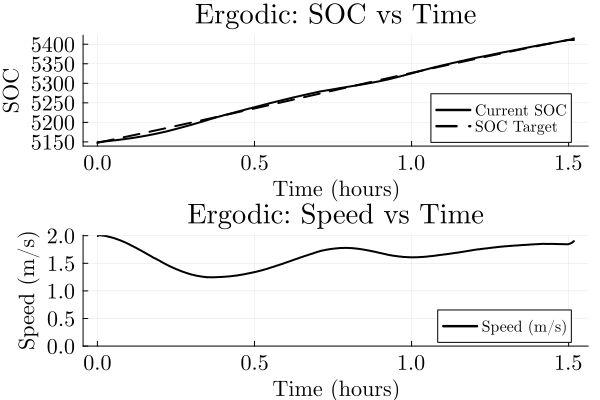}
    \caption{Ergodic mission: SOC and speed over a 90-minute variable-speed run. The controller adapts speed to track the nominal SOC trajectory.}\label{fig:90-variable-ergodic-soc-speed}
\end{figure}

\begin{figure}
    \centering
    \includegraphics[width=\columnwidth]{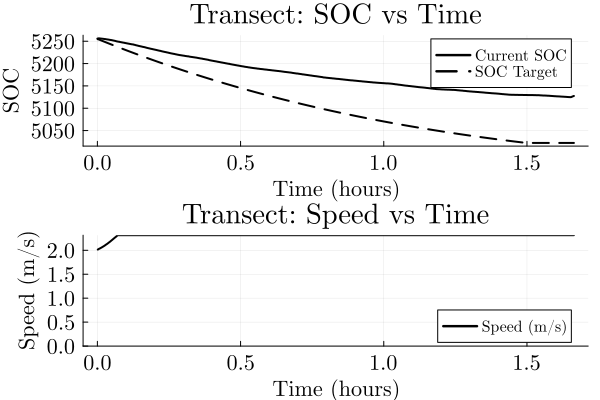}
    \caption{Transect mission SOC and speed over time for a 90-minute variable-speed mission. The controller saturates at maximum speed because the available solar resource exceeded forecasts, preventing tracking of the nominal SOC trajectory.}\label{fig:90-variable-transect-soc-speed}
\end{figure}

The trajectory in Fig.~\ref{fig:90-variable-path} begins to concentrate where the ASV measures wind speeds near the rated value. Fig.~\ref{fig:90-variable-measurement-error} shows the corresponding measured-minus-rated distributions. The ergodic run places more measurements near zero deviation than the transect run.

Figs.~\ref{fig:90-variable-ergodic-soc-speed} and~\ref{fig:90-variable-transect-soc-speed} show the combined SOC-and-speed plots discussed above. From these combined plots we observe that the SOC-tracking controller modulates the ASV's speed in response to deviations from the nominal SOC trajectory: when SOC falls below nominal the controller reduces speed to conserve energy, and when SOC exceeds nominal it increases speed to exploit available energy. The plots also show the speed approaching a steady value for the ergodic missions and similar SOC-tracking behavior for the transect missions. Finally, the combined plots make clear that the controller saturates at the maximum speed when the selected terminal SOC is lower than what the available solar resource can achieve during the 90-minute mission.

The need for active speed control becomes stronger in persistent missions, where the ASV must balance energy over multiple diurnal cycles. The 90-minute runs test the closed-loop implementation over the duration available here; they do not establish persistent energy balance.

\subsection{3 Hour Missions}\label{sec:experiment-three-hour}
To examine sustained performance over extended durations and observe early trends in energy-aware operation, we conducted a pair of 3-hour missions with closed-loop SOC control enabled.

\begin{figure}
    \centering
    \includegraphics[width=\columnwidth]{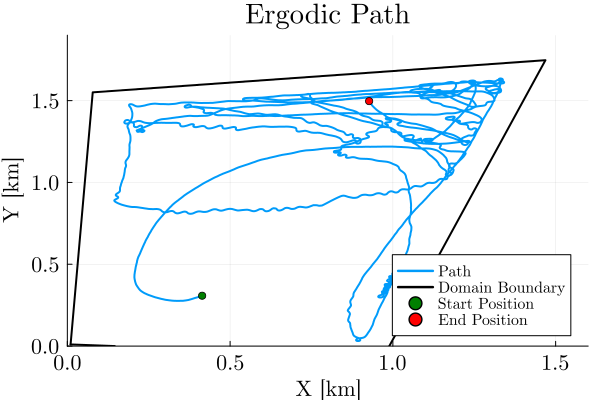}
    \caption{ASV trajectory for a 3-hour variable-speed ergodic mission. The path concentrates in the northeast region where wind speeds approach the ASV's rated value.}\label{fig:three-hour-trajectory}
\end{figure}

\begin{figure}
    \centering
    \includegraphics[width=\columnwidth]{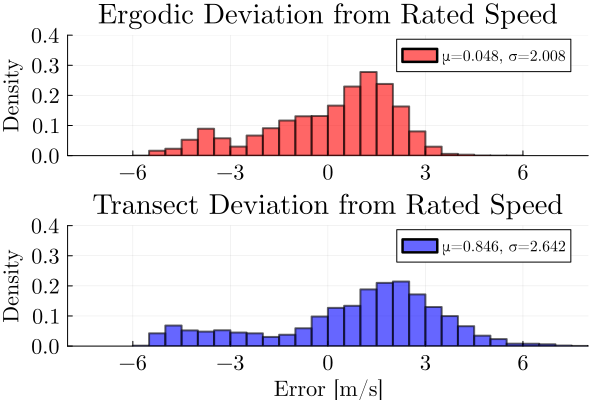}
    \caption{Measured-minus-rated wind-speed distributions for the 3-hour variable-speed missions. The mean deviations are $0.048\,\mathrm{m/s}$ for ergodic control and $0.846\,\mathrm{m/s}$ for the transect.}\label{fig:three-hour-measurement-error}
\end{figure}

The 3-hour ergodic trajectory in Fig.~\ref{fig:three-hour-trajectory} prioritizes the northern portion of the domain, where measured wind speeds are closer to the rated value. The measured-minus-rated histogram in Fig.~\ref{fig:three-hour-measurement-error} again shows a greater concentration near zero than the transect run.

\subsection{Summary of Results}\label{sec:experiment-summary}
\begin{table}[t]
    \centering
    \caption{Summary of measured-minus-rated wind-speed statistics for experimental missions.}
    \label{tab:measurement-errors}
    \begin{tabular}{l l r r}
        \toprule
        Mission type & Strategy & Mean (m/s) & SD (m/s) \\
        \midrule
        90 min, fixed speed & \textbf{Ergodic} & \textbf{0.081} & \textbf{2.146} \\
        90 min, fixed speed & Transect & 0.533 & 2.052 \\
        90 min, variable speed & \textbf{Ergodic} & \textbf{-0.894} & \textbf{1.595} \\
        90 min, variable speed & Transect & 1.176 & 2.183 \\
        3 hour, variable speed & \textbf{Ergodic} & \textbf{0.048} & \textbf{2.008} \\
        3 hour, variable speed & Transect & 0.846 & 2.642 \\
        \bottomrule
    \end{tabular}
\end{table}

The experiments support two conclusions. First, the ergodic planner concentrates measurements near the rated wind speed more strongly than the transect in these runs. Second, the SOC controller adjusts speed in response to the measured energy state while the adaptive planner continues to operate. 

Table~\ref{tab:measurement-errors} summarizes the measured-minus-rated wind-speed distributions. These deviations measure proximity to the scientific target; they are not errors relative to an independently measured ground-truth field. It is also worth noting that although comparisons between the ergodic adaptive and transect strategies were always performed on the same day, under comparable available solar irradiance, intra-day environmental variations still occur. Because of these variations in environmental conditions, the experiments do not establish improved global reconstruction or complete coverage of every interesting region. Rather, they demonstrate the practicality of implementing the targeted sampling and SOC-based execution together on the physical ASV.

\section{Conclusions and Future Work}\label{sec:conclusion}
We presented an energy-aware framework for adaptive exploration with a solar-powered autonomous surface vehicle. The framework uses clarity to represent evolving information, updates a target clarity map from in-situ wind measurements, and combines clarity-driven ergodic planning with SOC-based speed control. Simulations and field experiments at Jordan Lake demonstrate that these components operate together and direct sensing toward scientifically relevant conditions.

The main simulation result is that online adaptation increases sampling near the rated wind speed, but targeted sampling, spatial clarity, and full-field reconstruction are distinct objectives. Their balance depends on target selectivity, planner choice, and environmental structure. Mild selectivity can improve all three metrics when the interesting region is spatially coherent. Stronger selectivity and moving regions can instead trade reconstruction and spatial coverage for a greater concentration of relevant measurements. The field experiments demonstrate targeted sampling and SOC-based execution on the physical ASV.

The current target generator depends on the estimated mean and therefore may assign little value to an unvisited region that is incorrectly estimated to be uninteresting. Future work will investigate the use of additional running statistics, besides just the estimated mean, for target generation. We will also consider spatially varying kernel parameters for heterogeneous environments and couple the energy and information objectives directly within the online optimization.

\bibliographystyle{IEEEtran}
\bibliography{references}

\end{document}